\pdfoutput=1
\documentclass[11pt]{scaleai-paper}

\usepackage{amsmath}
\usepackage{amsfonts}
\usepackage{amssymb}
\usepackage{booktabs}
\usepackage{multirow}
\usepackage{subcaption}
\usepackage[square,numbers,sort&compress]{natbib}
\usepackage{xspace}
\usepackage{url}
\usepackage[colorlinks=true,linkcolor=scaleLink,citecolor=scaleLink,urlcolor=scaleLink]{hyperref}
\usepackage[capitalise,nameinlink]{cleveref}

\let\svthefootnote\thefootnote
\newcommand\freefootnote[1]{%
  \let\thefootnote\relax%
  \footnotetext{#1}%
  \let\thefootnote\svthefootnote%
}

\graphicspath{{figures/}}

\newcommand{\method}{OPSD\xspace}%
\definecolor{takeawayborder}{HTML}{C9CDD4}
\definecolor{takeawaybackground}{HTML}{F5F6F8}
\newcommand{\takeaway}[1]{%
  \par\smallskip
  \noindent\fcolorbox{takeawayborder}{takeawaybackground}{%
    \parbox{\dimexpr\linewidth-2\fboxsep-2\fboxrule\relax}{%
      \textbf{Takeaway.} #1}}%
  \par\vspace{1pt}
}
\definecolor{recommendationborder}{HTML}{6E8FB8}
\definecolor{recommendationbackground}{HTML}{EDF4FC}

\papertype{Scale AI Research}
\contact{Code and data: \url{https://github.com/zkjzou/PI-OPSD}}

\title{Predicting Privileged Information Effectiveness for On-Policy Self-Distillation}

\author[1,$\dagger$,$*$]{Kaijian~Zou}
\author[2]{Emiliano~Penaloza}
\author[3]{MohammadHossein~Rezaei}
\author[3]{Anisha~Gunjal}
\author[3]{Mohit~Raghavendra}
\author[1]{Lu~Wang}
\author[3]{Yunzhong~He}
\author[3,$*$]{Anas~Mahmoud}

\affil[1]{University of Michigan -- Ann Arbor}
\affil[2]{Mila}
\affil[3]{Scale AI}

\begin{document}

\freefootnote{${}^*$Corresponding authors: \href{mailto:zkjzou@umich.edu}{\texttt{zkjzou@umich.edu}},
\href{mailto:anas.mahmoud@scale.com}{\texttt{anas.mahmoud@scale.com}}.}
\freefootnote{${}^\dagger$Work done during an internship at Scale AI.}

\maketitle

\begin{abstract}
On-policy self-distillation (\textbf{OPSD}) enables large language models (LLMs) to
learn from privileged information (\textbf{PI}) without a separate teacher
model. However, prior work largely treats PI as a fixed design choice, leaving
the effects of its content and construction poorly understood. This challenge
is particularly important in multi-turn agentic settings, where trajectories
contain actions, observations, and reasoning that can be transformed into PI
in many ways. We systematically evaluate 16 PI configurations across three
multi-turn agentic benchmarks and three student models that vary in scale and
reasoning mode. We find that PI effectiveness depends on its content,
generation source, and the student model. Information-dense trajectories and
PI produced by more capable generators generally yield larger gains, whereas
thinking models benefit less consistently. To assess PI before training, we
compare PI utility with two KL-based metrics: sampled-token reverse KL ($D$) and
error-step KL contrast ($\Delta d^{-}_{\mathrm{err}}$). The KL-based metrics track post-training performance
more consistently than PI utility, which can be distorted by information
leakage. Empirically, selecting each task's PI by $\Delta d^{-}_{\mathrm{err}}$ also
improves post-training performance over random selection, most when the
PI candidates differ widely. Finally, effective PI remains beneficial when combined with GRPO,
allowing the combined objective to outperform a strong GRPO baseline. Overall,
our results provide practical guidance for constructing and evaluating PI
in OPSD training.

\end{abstract}

\section{Introduction}
\label{sec:introduction}

On-policy self-distillation (OPSD) enables LLMs to iteratively improve by turning successful solutions, environment feedback, and
experience from their own interactions into training signals without a
separate teacher model~\citep{penaloza2026privileged,zhao2026selfdistilled,
hubotter2026reinforcement,shenfeld2026selfdistillation}. Related
self-distillation methods have
already been used to train coding agents such as Composer~2.5 and agents for
long-horizon professional tasks~\citep{cursor2026composer,kale2026sdpo}. In
OPSD, the model learns from its own trajectories by matching the same model
conditioned on PI, thereby internalizing information unavailable at inference
time.

The effectiveness of OPSD, however, depends on what the teacher is allowed to
see. In general, PI can take many forms, including ground-truth
answers~\citep{zhao2026selfdistilled}, demonstrations from prior
tasks~\citep{shenfeld2026selfdistillation}, environment
feedback~\citep{hubotter2026reinforcement}, actions from successful
trajectories~\citep{penaloza2026privileged}, or hints derived from gold
solutions or trajectories, such as summaries~\citep{penaloza2026privileged},
skills~\citep{wang2026skillsd}, checklists, structured criteria such as
rubrics~\citep{rezaei2026rubricguided,gu2026rubricconditioned}, and targeted
feedback~\citep{yeo2026hintsd,cursor2026composer}. These choices
determine both the information available to the
teacher and how that information changes its distribution relative to the
student. However, especially on multi-turn agentic tasks, existing studies evaluate only a
limited range of PI formats across a small set of models and tasks
\citep{lu2026sdar,yang2026selfdistilledrlvr,wang2026skillsd,
yeo2026hintsd,yang2026opid}. It remains unclear what makes PI
produce useful learning signals for the student policy, whether the same choices
work across agentic domains and model types, and, critically, whether PI quality
can be assessed before training. Reliable offline metrics would allow candidate
PI formats and sources to be compared before training, avoiding a separate
costly OPSD run for every candidate and focusing compute on
the~most~promising~PI.

\begin{figure}[t]
  \centering
  \includegraphics[angle=270,width=0.9\linewidth,trim=198 198 198 198,clip]{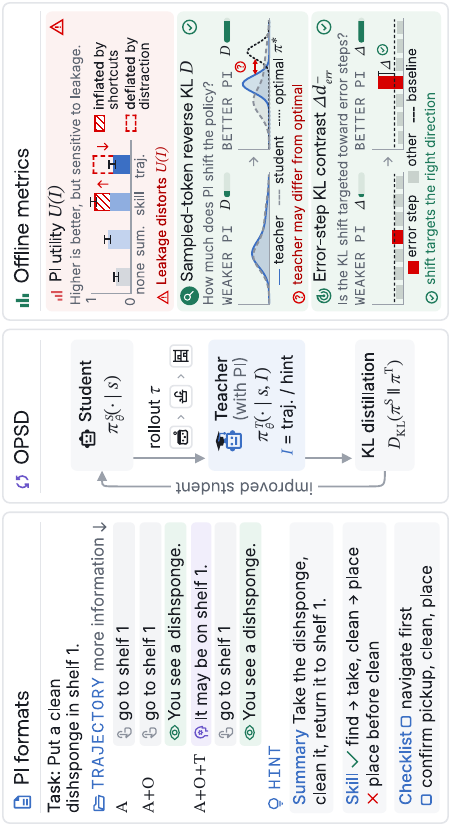}
  \caption{Overview of the PI formats and offline metrics studied.
  \textbf{Left---PI formats:} reference trajectories retain actions
  (\textbf{A}), add observations (\textbf{A+O}) or thoughts (\textbf{A+T}),
  or include both (\textbf{A+O+T}). Compressed hints include a plain \textbf{Summary}, a
  \textbf{Skill} with mistake analysis and reusable strategies, and a
  \textbf{Checklist} of task requirements derived from reference and failed
  student trajectories. \textbf{Right---offline metrics:} \textbf{PI utility}
  measures the teacher's reward gain from PI, but leakage can inflate or deflate
  it. \textbf{Sampled-token reverse KL} $D$ measures shift magnitude; better PI
  is generally associated with a larger $D$, but $D$ does not indicate whether
  the shift is beneficial. \textbf{Error-step KL contrast}
  $\Delta d^{-}_{\mathrm{err}}$ measures whether that shift targets the
  identified errors.}
  \label{fig:overview}
\end{figure}

We study these questions through a systematic evaluation of OPSD on three
multi-turn agentic benchmarks: ALFWorld~\citep{shridhar2021alfworld},
AppWorld~\citep{trivedi2024appworld}, and
$\tau^2$-retail~\citep{barres2025tau2bench}.
Figure~\ref{fig:overview} summarizes the PI formats and offline metrics
evaluated in our study. We cover 16 PI configurations spanning reference
trajectories and self- and expert-generated hints, including summaries, skills,
and checklists. We train
Qwen3-4B-Instruct, Qwen3-4B-Thinking, and Qwen3-30B-A3B-Instruct
\citep{yang2025qwen3}, allowing us
to compare PI effects across instruction and thinking models and across model
scales. Beyond downstream performance, we evaluate three offline metrics:
\textbf{PI utility}~\citep{penaloza2026privileged} compares teacher reward with
and without PI, but can be distorted by information leakage;
\textbf{sampled-token reverse KL} $D$ measures the PI-induced probability shift
on student-sampled tokens; and \textbf{error-step KL contrast}
$\Delta d^{-}_{\mathrm{err}}$ measures whether PI suppresses tokens more at
identified errors than at matched steps from the same task. Together, these
experiments ask not only which PI works, but also what distinguishes useful PI
from information that merely makes the teacher more capable when PI is
present.

Our results show that PI effectiveness depends on its content, source, and
the student model. PI formats with information-dense trajectories usually outperform compressed
hints, and adding environment observations consistently improves the
Instruct students on ALFWorld and AppWorld. Additional teacher reasoning is
less reliable and can reduce the gains of the thinking models. Among hint
formats, skills are generally strongest, suggesting that additional guidance,
such as an analysis of mistakes and reusable strategies, is more effective
than a plain trajectory summary. More capable generators also tend to produce
more effective PI, particularly for hint formats, showing that PI quality
depends on both its format and source.

Among the three offline metrics, error-step KL contrast better distinguishes useful PI and improves per-task selection when PI candidates vary in source or strength. Sampled-token reverse KL tracks performance in eight of nine settings but does not capture the direction of the shift, while PI utility is unreliable because of information leakage (Figure~\ref{fig:metrics_comparison}). Our controlled evaluations on ALFWorld also show that PI utility can be distorted by information leakage and that reverse KL can assign large values to unhelpful shifts. Using the most effective PI format, we further show that combining OPSD with GRPO can outperform a strong GRPO baseline.

Our main contributions are:
\begin{itemize}
  \item \textbf{Systematic study of PI for OPSD.} To our knowledge, this is the
  first controlled study of PI construction for OPSD, spanning 16 PI
  configurations and three student models across three multi-turn agentic
  benchmarks.
  \item \textbf{Practical PI design recommendations.} We identify how PI
  content, format, generation source, and student-model choice shape OPSD gains.
  \item \textbf{KL-based PI assessment and selection.} We propose two offline KL
  diagnostics, evaluate what they measure through controlled studies, and show
  that error-step KL contrast can guide per-task PI selection before costly OPSD
  training.
  \item \textbf{Combining OPSD with GRPO.} We show that the most effective PI
  enables combined OPSD and GRPO training to outperform matched GRPO baselines.
\end{itemize}

\section{Analyzing Privileged Information for OPSD}
\label{sec:opsd_pi_analysis}

\subsection{On-Policy Self-Distillation and PI Construction}
\label{sec:opsd_pi_construction}
On-policy distillation (OPD) transfers knowledge from a teacher on trajectories
generated by the student, providing token-level supervision on the states that
the student actually visits~\citep{agarwal2024onpolicy,
yang2026generalizedopd}. OPSD instantiates this approach without a separate
teacher model: the student and teacher share parameters, but only the teacher
observes PI. Let $\theta$ denote their shared parameters, $I$ the PI, and
$\pi_\theta^S$ and $\pi_\theta^T$ the student and PI-conditioned teacher
policies. For a student trajectory $\tau=(y_1,\ldots,y_{L_\tau})$, $s_t$ is the
context before token $y_t$, and $L_\tau$ is the trajectory length. OPSD
minimizes the reverse KL from the student to the PI-conditioned teacher:
\begin{equation}
  \mathcal{L}_{\mathrm{OPSD}}(\theta)
  = \mathbb{E}_{\tau\sim\pi_\theta^S}
    \left[\frac{1}{L_\tau}\sum_{t=1}^{L_\tau}
    D_{\mathrm{KL}}\!\left(
      \pi_\theta^S(\cdot\mid s_t)
      \,\|\,
      \operatorname{sg}\!\left[\pi_\theta^T(\cdot\mid s_t,I)\right]
    \right)\right].
  \label{eq:opsd_objective}
\end{equation}
Here, $D_{\mathrm{KL}}$ compares the two next-token distributions, and
$\operatorname{sg}$ stops gradients through the teacher.

This formulation has been applied with ground-truth
answers~\citep{zhao2026selfdistilled}, reflective environment
feedback~\citep{hubotter2026reinforcement}, and information retained across
continual-learning tasks~\citep{shenfeld2026selfdistillation}. Because the two
policies differ only in access to PI, PI construction directly determines the
token-level training signal~\citep{penaloza2026privileged}.

We study seven PI formats constructed using GLM-5.2 and Qwen3.7-Plus. The
formats comprise four trajectory variants and three types of hints.
Figure~\ref{fig:overview} illustrates these formats and the overall OPSD
pipeline.
Appendix~\ref{app:pi_construction} outlines the PI construction protocol.

\paragraph{Trajectories.}
Following \citet{penaloza2026privileged}, we sample successful reference
trajectories from the expert models and retain different components: actions
only (\emph{Action}), actions with environment observations
(\emph{Action $+$ Obs.}), actions with reasoning (\emph{Action $+$ Thou.}), or
actions, observations, and reasoning (\emph{Action $+$ Obs.\ $+$ Thou.}). These variants isolate whether effective PI
requires only the expert's actions, the environmental context surrounding
those actions, or the reasoning used to derive them.

\paragraph{Summary and Skill Hints.}
Summary hints condense a reference trajectory into a brief description of its
solution~\citep{penaloza2026privileged}. Skill hints provide more detailed
guidance by combining high-level strategies, mistake analysis, and reusable
skills derived from the reference trajectory~\citep{wang2026skillsd}. We also
use the student model itself to generate both types of hints.

\paragraph{Checklist Hints.}
We generate checklists from a successful reference and a failed student
trajectory, listing task requirements for the teacher. Like rubric-conditioned
self-distillation~\citep{rezaei2026rubricguided,gu2026rubricconditioned}, this
format provides structured criteria rather than a target response.

\subsection{Offline Metrics}

\label{sec:pi_metrics}
Training a separate student for every PI candidate is computationally
expensive. We therefore study offline metrics that can assess PI effectiveness
before training and help identify what distinguishes PI that produces better learning signals for the student policy.

Using the notation above, let $R(\tau)\in\{0,1\}$ denote whether trajectory
$\tau$ completes its task successfully. We define the following metrics:

\paragraph{PI utility.}
PI utility measures the improvement in task success obtained by exposing the
same model to PI:
\begin{equation}
    U(I)
    = \mathbb{E}_{\tau\sim\pi_\theta^T(\cdot\mid I)}[R(\tau)]
    - \mathbb{E}_{\tau\sim\pi_\theta^S}[R(\tau)].
    \label{eq:pi_utility}
\end{equation}
A positive value indicates that the teacher solves more tasks when conditioned
on PI, motivating the hypothesis that higher PI utility leads to more effective
OPSD training.

\paragraph{Sampled-token Reverse KL.}
For a student trajectory $\tau=(y_1,\ldots,y_{L_\tau})$, we estimate the reverse KL
from the sampled-token log-ratio, using its absolute magnitude so that shifts
in either direction count:
\begin{equation}
    d_I(\tau)
    = \frac{1}{L_\tau}\sum_{t=1}^{L_\tau}
      \left|\log\frac{\pi_\theta^S(y_t\mid s_t)}
                     {\pi_\theta^T(y_t\mid s_t,I)}\right|,
    \qquad y_t\sim\pi_\theta^S(\cdot\mid s_t).
    \label{eq:sampled_reverse_kl}
\end{equation}
Intuitively, a larger reverse KL indicates that PI more strongly changes the
teacher's predictions on student-visited states, providing a stronger potential
learning signal. We report $D(I)=\mathbb{E}_{\tau\sim\pi_\theta^S}[d_I(\tau)]$
over the sampled student trajectories.

\paragraph{Error-step KL Contrast ($\Delta d^{-}_{\mathrm{err}}$).}
This metric measures whether PI suppresses tokens more strongly at identified
errors than at comparable steps in other rollouts. An agent trajectory contains
multiple interaction steps, each of which may contain several generated tokens.
For step $j$, let $\mathcal{T}_j(\tau)\subseteq\{1,\ldots,L_\tau\}$ denote its token
indices and define
\begin{equation}
  d^-_I(\tau,j)
  = \frac{1}{|\mathcal{T}_j(\tau)|}
    \sum_{t\in\mathcal{T}_j(\tau)}
    \left[\log\frac{\pi_\theta^S(y_t\mid s_t)}
                       {\pi_\theta^T(y_t\mid s_t,I)}\right]_+,
  \qquad [z]_+=\max(z,0).
  \label{eq:error_step_suppression}
\end{equation}
Because KL can vary with step position, we
subtract a task- and position-matched baseline to reduce these effects. For
each PI condition $I$, task $x$, and absolute step index $j$, the baseline is
\begin{equation}
  b_I(x,j)
  = \frac{1}{|\mathcal{R}_{x,j}|}
    \sum_{\tau'\in\mathcal{R}_{x,j}} d^-_I(\tau',j),
  \label{eq:error_step_baseline}
\end{equation}
where $\mathcal{R}_{x,j}$ is the set of sampled trajectories for task $x$ that
reach step $j$. For each trajectory, a judge labels the set $E_\tau$ of error
step indices. With
$\mathcal{E}_F=\{(\tau,j):R(\tau)=0,\ j\in E_\tau\}$, we report
\begin{equation}
  \Delta d^{-}_{\mathrm{err}}(I)
  = \frac{1}{|\mathcal{E}_F|}
    \sum_{(\tau,j)\in\mathcal{E}_F}
    \left[d^-_I(\tau,j)-b_I(x_\tau,j)\right],
  \label{eq:delta_error}
\end{equation}
where $\mathcal{E}_F$ contains every flagged step from a failed trajectory,
$x_\tau$ is the task associated with $\tau$, and each flagged step receives
equal weight. Positive values indicate that PI targets identified errors more
strongly than matched positions within the same task.

\subsection{Experimental Setup}
\label{sec:experimental_setup}
We evaluate OPSD on ALFWorld~\citep{shridhar2021alfworld},
AppWorld~\citep{trivedi2024appworld}, and
$\tau^2$-retail~\citep{barres2025tau2bench}, using
Qwen3-4B-Instruct-2507, Qwen3-4B-Thinking-2507, and
Qwen3-30B-A3B-Instruct-2507 as students. We implement OPSD in
SkyRL~\citep{cao2025skyrl} with the sampled-token reverse-KL objective and
report pass@1 from the best
three-checkpoint window.

Offline metrics are computed on training or development tasks disjoint from
the evaluation sets used to measure post-training performance.
Appendices~\ref{app:details} and~\ref{app:diagnostics} provide the benchmark
splits, sampling protocols, training hyperparameters, and evaluation settings.

\section{Results and Analysis}
\label{sec:experiments}

\subsection{Which PI Formats Work Best?}
\label{sec:main_results}

\begin{figure}[t]
  \centering
  \begin{minipage}{0.9\linewidth}
  \centering
  \begin{subfigure}[t]{0.32\linewidth}
    \includegraphics[width=\linewidth]{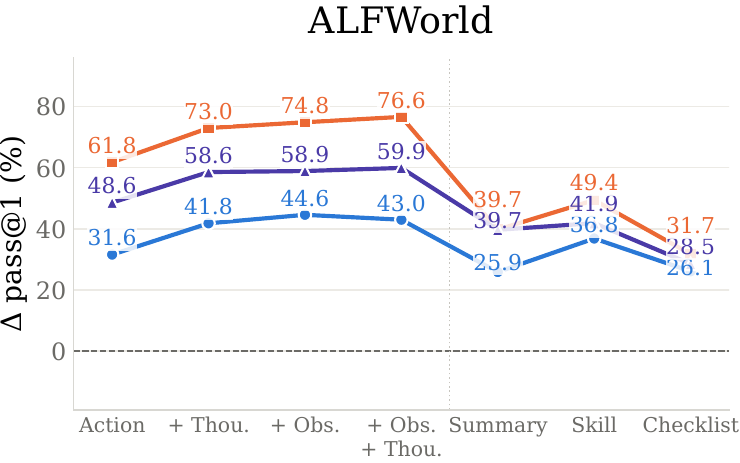}
    \caption{}
    \label{fig:format_alfworld}
  \end{subfigure}\hfill
  \begin{subfigure}[t]{0.32\linewidth}
    \includegraphics[width=\linewidth]{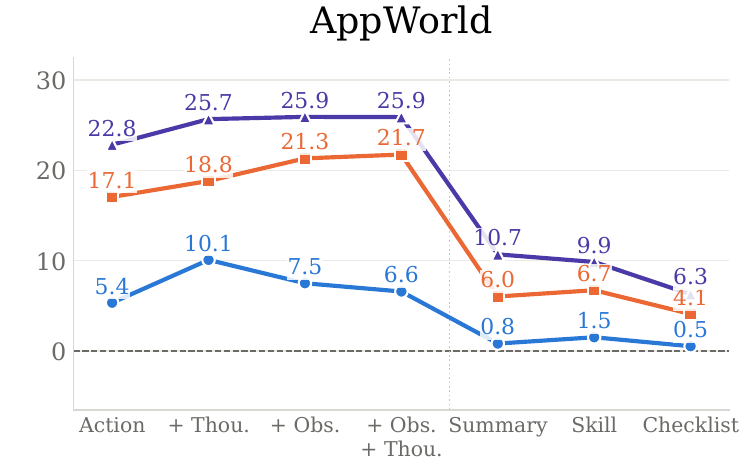}
    \caption{}
    \label{fig:format_appworld}
  \end{subfigure}\hfill
  \begin{subfigure}[t]{0.32\linewidth}
    \includegraphics[width=\linewidth]{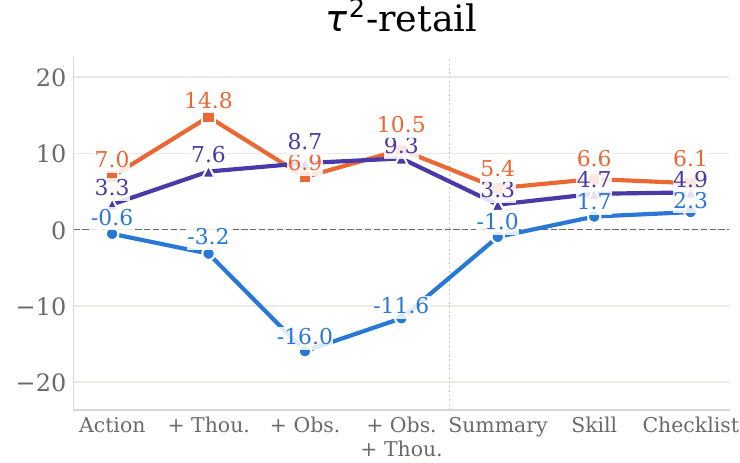}
    \caption{}
    \label{fig:format_tau2}
  \end{subfigure}

  \begin{subfigure}[t]{0.32\linewidth}
    \includegraphics[width=\linewidth]{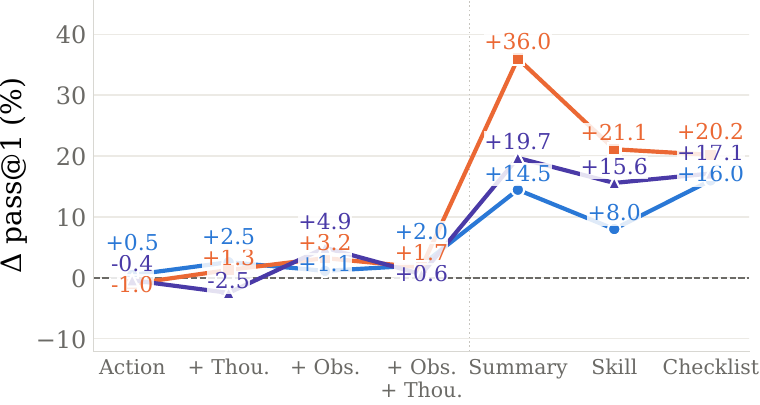}
    \caption{}
    \label{fig:source_alfworld}
  \end{subfigure}\hfill
  \begin{subfigure}[t]{0.32\linewidth}
    \includegraphics[width=\linewidth]{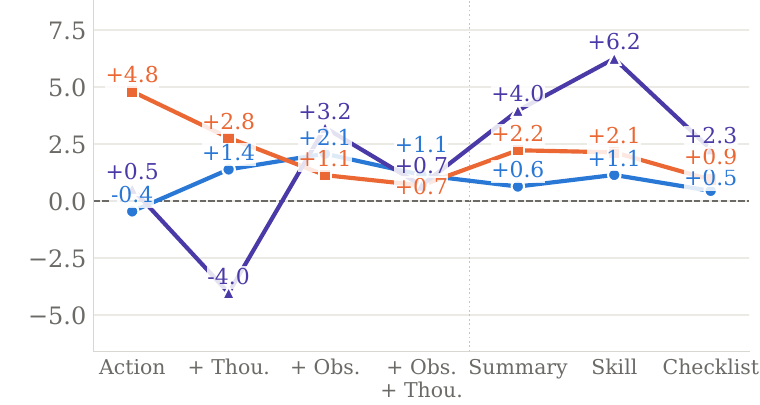}
    \caption{}
    \label{fig:source_appworld}
  \end{subfigure}\hfill
  \begin{subfigure}[t]{0.32\linewidth}
    \includegraphics[width=\linewidth]{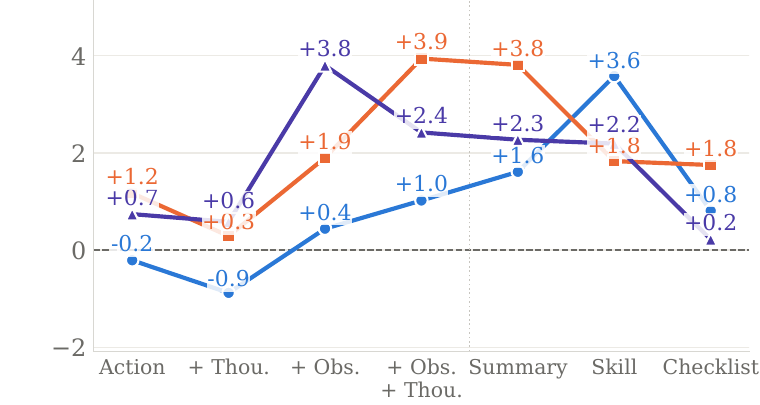}
    \caption{}
    \label{fig:source_tau2}
  \end{subfigure}

  \begin{subfigure}[t]{0.32\linewidth}
    \includegraphics[width=\linewidth]{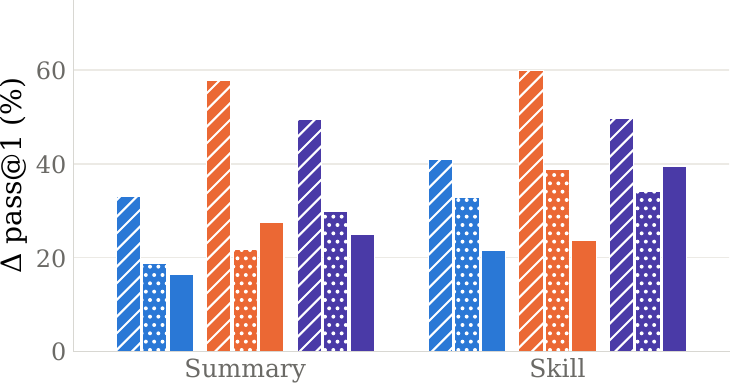}
    \caption{}
    \label{fig:self_hint_alfworld}
  \end{subfigure}\hfill
  \begin{subfigure}[t]{0.32\linewidth}
    \includegraphics[width=\linewidth]{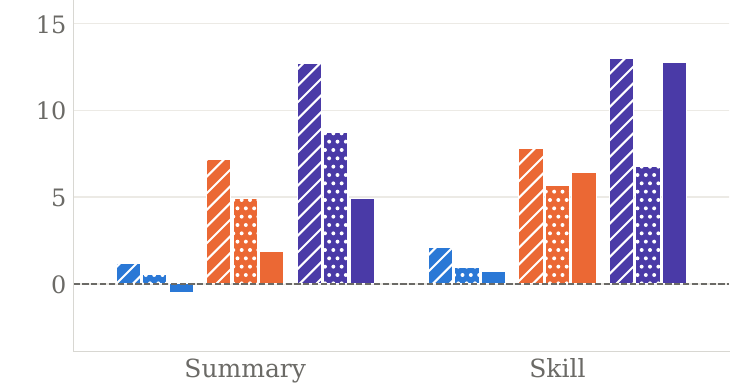}
    \caption{}
    \label{fig:self_hint_appworld}
  \end{subfigure}\hfill
  \begin{subfigure}[t]{0.32\linewidth}
    \includegraphics[width=\linewidth]{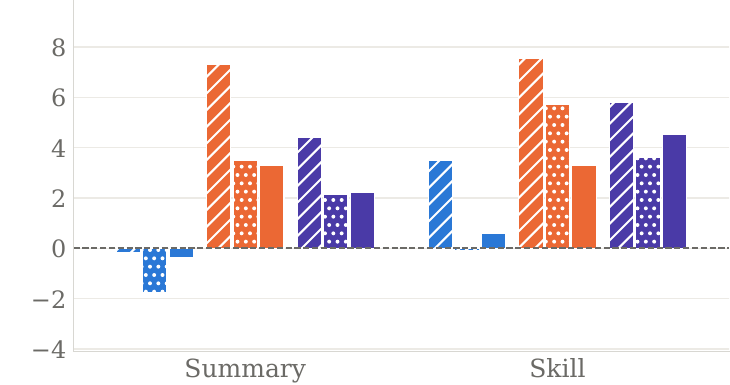}
    \caption{}
    \label{fig:self_hint_tau2}
  \end{subfigure}

  \includegraphics[width=0.62\linewidth]{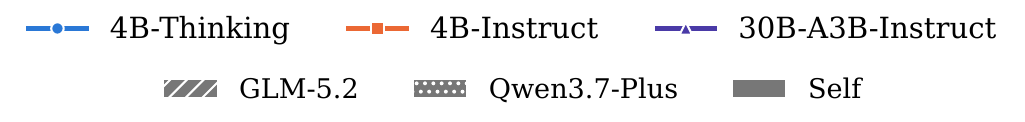}
  \end{minipage}
  \caption{OPSD improvement relative to baseline across PI formats and sources;
  columns correspond to ALFWorld, AppWorld, and $\tau^2$-retail.
  \textbf{(a)--(c)} Mean performance across GLM and Qwen sources.
  \textbf{(d)--(f)} GLM minus Qwen pass@1; positive values favor GLM.
  \textbf{(g)--(i)} Summary and Skill gains for GLM, Qwen, and Self.}
  \label{fig:format_comparison}
\end{figure}

Figure~\ref{fig:format_comparison} compares PI formats (top), generation
sources (middle), and self-generated hints (bottom) across three benchmarks and
student models.

\paragraph{Trajectory formats generally outperform hints, but detailed PI can
hurt already-capable models.}
Trajectory PI generally outperforms compressed hints
(Figures~\ref{fig:format_alfworld}--\ref{fig:format_tau2}). Instruct models
benefit further when trajectories include observations and teacher thinking
tokens, suggesting that additional context strengthens the training signal.
However, richer PI is not universally beneficial for the thinking model:
adding teacher thinking tokens to Action $+$ Obs.\ hurts its performance on ALFWorld and AppWorld,
while on $\tau^2$, adding either observations or thinking tokens causes even
larger declines in Figure~\ref{fig:format_tau2}. Its
high pass@4 of $88.9$ on $\tau^2$ suggests that it already reasons
effectively, leaving additional trajectory context less room to help and more
potential to interfere with its existing strategy.

\paragraph{Effective PI provides context beyond the correct solution.}
Skills, which augment the solution with mistake analysis and reusable
strategies, are often the strongest hint format. They lead for all three
students on ALFWorld, both 4B students on AppWorld, and 4B-Instruct on
$\tau^2$. Together with the gains from adding environment observations to
trajectories, these results suggest
that effective PI should do more than reveal or summarize a correct solution:
it should provide context that helps the student understand its mistakes and
how to act successfully in the environment.

\paragraph{The PI source matters, especially when generating hints.}
Figures~\ref{fig:source_alfworld}--\ref{fig:source_tau2} show that GLM-5.2 PI
almost always outperforms Qwen3.7-Plus PI: for
4B-Instruct, GLM wins 20 of 21 comparisons, and its lead reaches 36.0 points on
ALFWorld hints but stays within 3.2 points for trajectory formats. Source choice
thus matters most when trajectories are rewritten into guidance: for trajectory
PI, the source determines only the reference behavior the teacher sees, whereas
for hints the generator also decides which information is kept and how it is
expressed.

\paragraph{Self-generated hints yield positive performance gains.}
For 4B-Instruct, both self-generated Summary and Skill hints improve
performance on all three benchmarks
(Figures~\ref{fig:self_hint_alfworld}--\ref{fig:self_hint_tau2}). These results suggest that the student can act as its own hint generator given the reference solution and still benefit from OPSD.

\subsection{What Do PI Metrics Reveal?}
\label{sec:metric_observations}
\begin{figure}[t]
  \centering
  \begin{minipage}{0.9\linewidth}
  \centering
  \begin{subfigure}[t]{0.32\linewidth}
    \includegraphics[width=\linewidth]{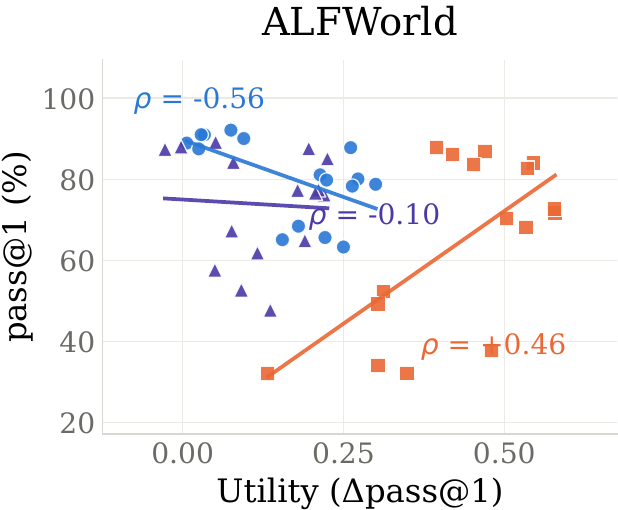}
    \caption{}
    \label{fig:utility_alfworld}
  \end{subfigure}\hfill
  \begin{subfigure}[t]{0.32\linewidth}
    \includegraphics[width=\linewidth]{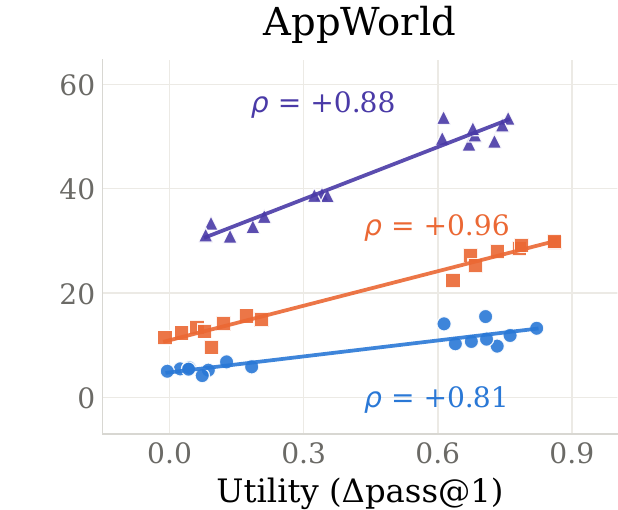}
    \caption{}
    \label{fig:utility_appworld}
  \end{subfigure}\hfill
  \begin{subfigure}[t]{0.32\linewidth}
    \includegraphics[width=\linewidth]{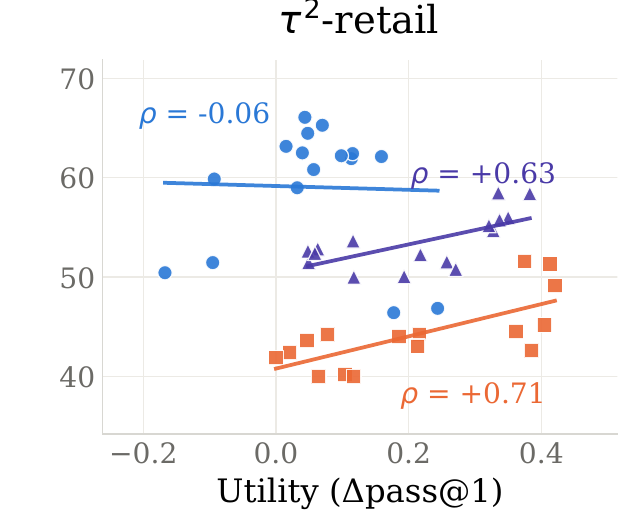}
    \caption{}
    \label{fig:utility_tau2}
  \end{subfigure}

  \begin{subfigure}[t]{0.32\linewidth}
    \includegraphics[width=\linewidth]{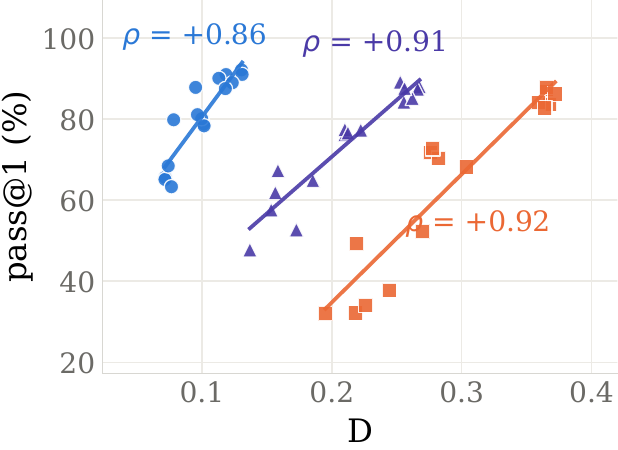}
    \caption{}
    \label{fig:reverse_kl_alfworld}
  \end{subfigure}\hfill
  \begin{subfigure}[t]{0.32\linewidth}
    \includegraphics[width=\linewidth]{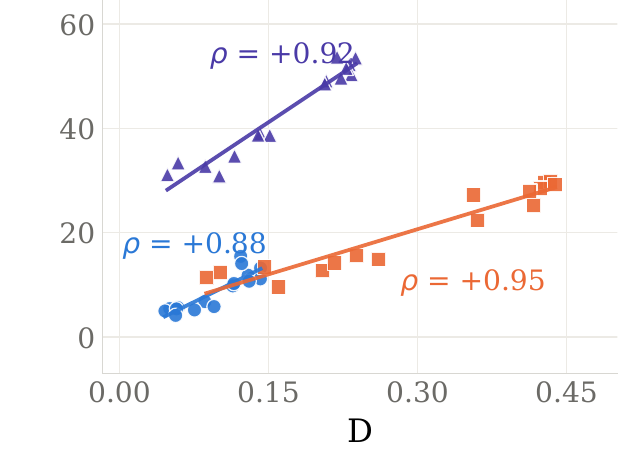}
    \caption{}
    \label{fig:reverse_kl_appworld}
  \end{subfigure}\hfill
  \begin{subfigure}[t]{0.32\linewidth}
    \includegraphics[width=\linewidth]{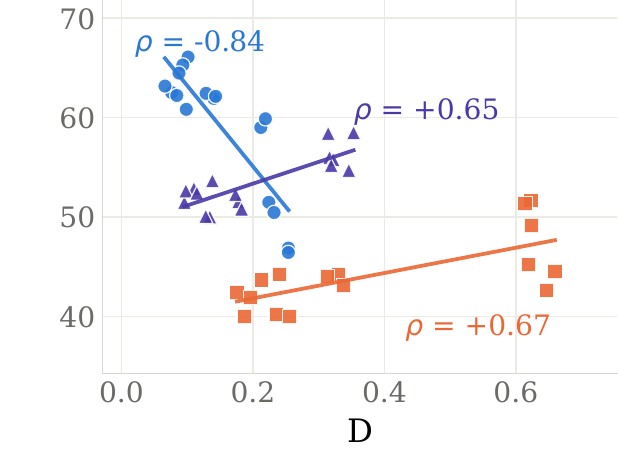}
    \caption{}
    \label{fig:reverse_kl_tau2}
  \end{subfigure}

  \begin{subfigure}[t]{0.32\linewidth}
    \includegraphics[width=\linewidth]{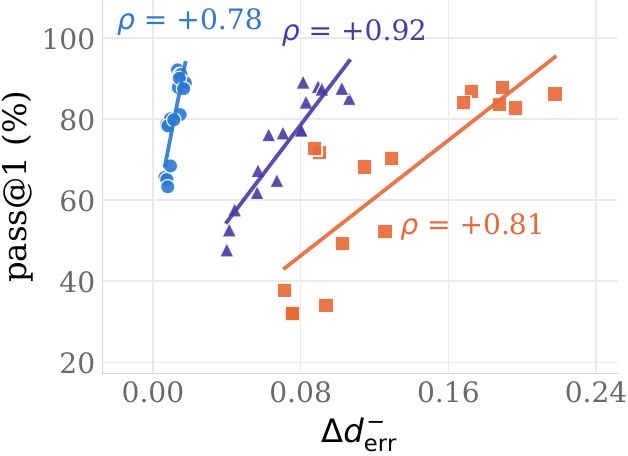}
    \caption{}
    \label{fig:error_contrast_alfworld}
  \end{subfigure}\hfill
  \begin{subfigure}[t]{0.32\linewidth}
    \includegraphics[width=\linewidth]{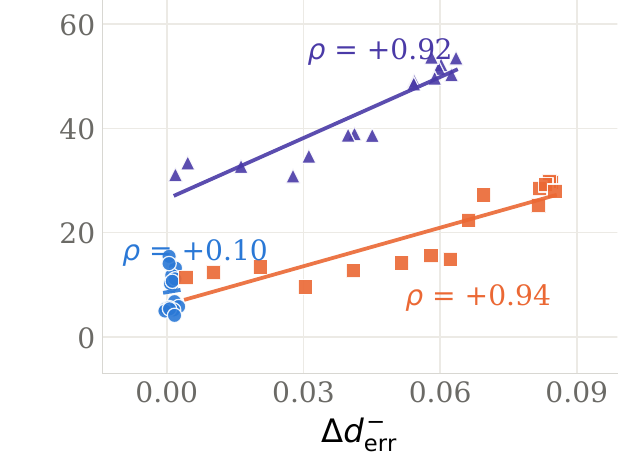}
    \caption{}
    \label{fig:error_contrast_appworld}
  \end{subfigure}\hfill
  \begin{subfigure}[t]{0.32\linewidth}
    \includegraphics[width=\linewidth]{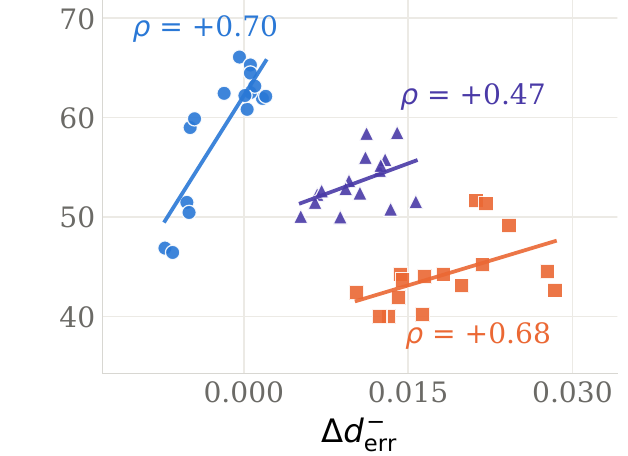}
    \caption{}
    \label{fig:error_contrast_tau2}
  \end{subfigure}

  \includegraphics[width=0.55\linewidth]{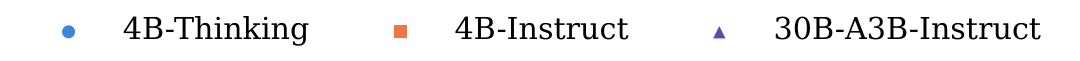}
  \end{minipage}
  \caption{Offline PI metrics versus post-training OPSD performance; columns
  correspond to ALFWorld, AppWorld, and $\tau^2$-retail. Offline metrics
  are computed on training or development tasks for all three benchmarks.
  \textbf{(a)--(c)} PI utility. \textbf{(d)--(f)} Sampled-token reverse KL
  $D$. \textbf{(g)--(i)} Error-step KL contrast
  $\Delta d^{-}_{\mathrm{err}}$. Lines show per-model linear fits; $\rho$
  denotes Spearman correlation.}
  \label{fig:metrics_comparison}
\end{figure}

Given the variation across PI configurations above and the substantial cost of
training each one, we next ask whether post-training effectiveness can be
assessed without running OPSD for every candidate.
Figure~\ref{fig:metrics_comparison} plots each PI metric against post-training
performance for every student under each PI configuration across ALFWorld,
AppWorld, and $\tau^2$-retail. All diagnostics are computed on training or
development tasks disjoint from the post-training evaluation sets. These
comparisons show how utility, the
distributional-shift statistic $D$, and the error-step contrast
$\Delta d^{-}_{\mathrm{err}}$ relate to the performance gains produced by
different PI configurations.

\paragraph{PI utility can be inflated by information leakage.}
The PI utility measures the improvement in the success rate of a model when conditioned on a given PI. We find that this metric can reward PI that reveals
answer-specific information unavailable to the student,
rather than PI that provides a transferable training signal. On AppWorld,
the utility metric correlates strongly with post-training performance for all three
students ($\rho=0.81$--$0.96$). The relationship is less consistent on
ALFWorld, where the correlation is positive for 4B-Instruct ($\rho=0.46$) but
negative for 4B-Thinking and 30B-Instruct ($\rho=-0.56$ and $-0.10$), and on
$\tau^2$, where correlations range from $-0.06$ to $0.71$. On ALFWorld, the
PI-conditioned model can copy the reference trajectory's final action without
reproducing the preceding interaction sequence. On $\tau^2$, it can extract an
answer from PI without executing the actions required to update the
environment's database (Appendix~\ref{app:utility_leakage}). In these cases,
utility measures the model's ability
to exploit leaked solution information rather than how well the PI transfers
through training.

\takeaway{PI utility is unreliable when information leakage changes evaluation
behavior without providing transferable guidance, so higher utility does not
imply larger OPSD gains.}

\paragraph{Larger distributional shifts usually accompany better performance.}
We next examine $D$, the mean absolute sampled-token reverse KL between the student and PI-conditioned teacher. It correlates positively with post-training
performance in eight of the nine student--benchmark combinations
($\rho=0.65$--$0.95$). The exception is 4B-Thinking on
$\tau^2$, where the correlation is strongly negative ($\rho=-0.84$).
Its pass@4 of $88.9$ suggests that a large shift can overwrite useful
behavior already present in the policy. Our PI ablations point the same way: the
student's own failed rollouts induce a larger $D$ than its successful ones, even
though they make far worse PI (Appendix~\ref{sec:corrupted_pi}). Thus, shift direction
and location matter~in~addition~to~magnitude.

\takeaway{A larger $D$ generally indicates a stronger PI training signal, but
does not reveal whether the induced shift improves the policy.}

\paragraph{Error-step KL contrast consistently tracks post-training performance.}
Across all nine student--benchmark combinations,
$\Delta d^{-}_{\mathrm{err}}$ is positively correlated with post-training
performance ($\rho=0.10$--$0.94$). A larger contrast means that PI more
strongly suppresses the student's sampled tokens at judge-identified error
steps than at comparable positions from the same task. Thus, PI formats that
concentrate the distributional shift at critical mistakes tend to produce
larger training gains. This relationship is particularly strong for
4B-Instruct on AppWorld ($\rho=0.94$) and 30B-Instruct on ALFWorld
($\rho=0.92$). On $\tau^2$, the contrast remains positively associated with
performance for 4B-Thinking ($\rho=0.70$), even though several trajectory
formats have negative values. These small or negative contrasts and limited
OPSD gains suggest poorly targeted supervision. This interpretation is
consistent with
\citet{kaur2026rethinking}, who show that privileged distillation can penalize
productive reconsideration tokens in thinking models: supervision may
insufficiently suppress errors while disrupting useful behavior.

\takeaway{Effective PI concentrates token-level suppression on actual mistakes
rather than comparable non-error steps, yielding a larger
positive~$\Delta d^{-}_{\mathrm{err}}$.}

\section{How Well Does OPSD Combine with GRPO?}
\label{sec:opsd_grpo}

\begin{figure}[t]
  \centering
  \includegraphics[width=0.9\textwidth]{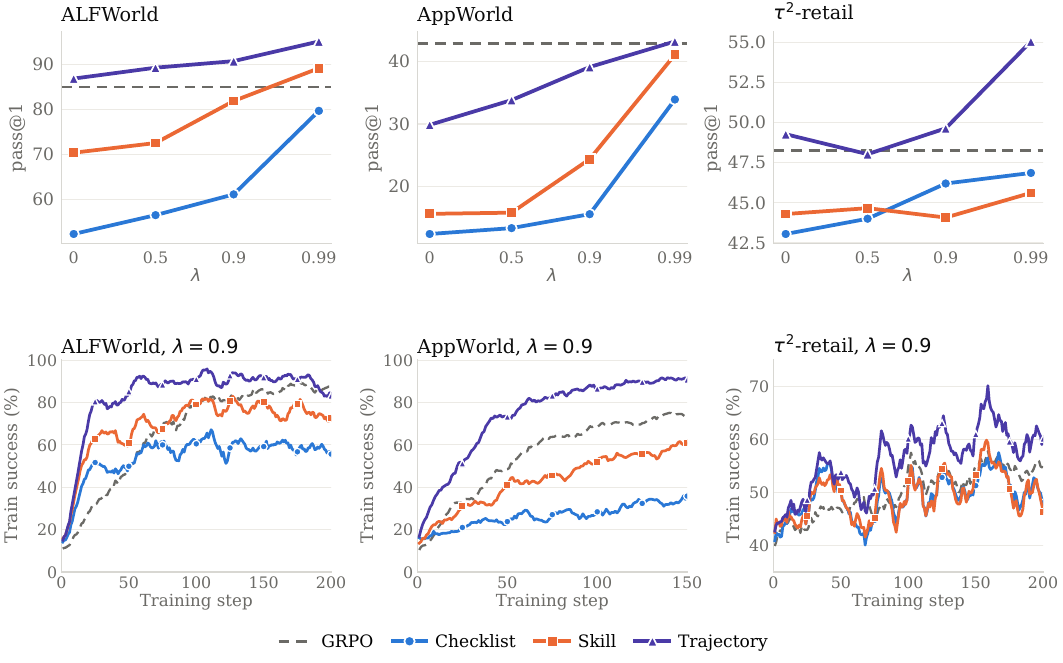}
  \caption{OPSD+GRPO performance on ALFWorld, AppWorld, and $\tau^2$-retail.
  The top row shows performance across objective weights, and the bottom row
  shows training curves at $\lambda=0.9$.
  Here, $\lambda=0$ is vanilla OPSD and $\lambda=1$ is GRPO; dashed lines show
  the matched GRPO baseline.}
  \label{fig:beta_sweep}
\end{figure}

\subsection{Combined Training Objective}
\label{sec:combined_objective}
Rather than optimizing over the many possible ways to combine token-level
distillation and GRPO outcome rewards~\citep{shao2024deepseekmath}, we use a
simple weighted objective to test
whether PI identified as effective under vanilla OPSD remains effective when
combined with GRPO. We use Qwen3-4B-Instruct-2507 for all experiments in this
section.
We combine the GRPO and OPSD signals at the advantage level:
\begin{equation}
  A^{\mathrm{OPSD+GRPO}}
  = \lambda A^{\mathrm{GRPO}}
  + (1-\lambda)A^{\mathrm{OPSD}},
  \label{eq:combined_objective}
\end{equation}
where $A^{\mathrm{GRPO}}$ is the outcome-based group-relative advantage,
$A^{\mathrm{OPSD}}$ is the token-level self-distillation advantage, and
$\lambda\in[0,1]$ controls their relative weight.

Figure~\ref{fig:beta_sweep} compares four values of $\lambda$ across three PI
formats with matched GRPO-only controls on ALFWorld, AppWorld, and $\tau^2$. The
$\lambda=0$ endpoint corresponds to vanilla OPSD.

\subsection{Effect of Objective Weighting}
\label{sec:objective_weighting}

\paragraph{PI quality persists: better PI is still better after adding GRPO.}
Trajectory PI performs best under vanilla OPSD and remains strongest across
most values of $\lambda$, followed by Skill and Checklist PI. This ordering is
consistent through $\lambda=0.99$ on all benchmarks. Thus, combining OPSD
with outcome-based rewards does not eliminate differences in PI quality:
stronger PI continues to provide a more effective auxiliary training signal.

\takeaway{The PI format that performs best under vanilla OPSD generally remains
the strongest after adding GRPO, so outcome rewards complement rather than
replace effective PI selection.}

\paragraph{GRPO strengthens OPSD.}
Increasing $\lambda$ generally improves performance for all three PI formats,
with the largest gains occurring for the weaker Checklist and Skill formats on ALFWorld and AppWorld.
On ALFWorld, several Skill and Trajectory configurations exceed the
GRPO-only baseline score, with Trajectory PI reaching $95.13$ at $\lambda=0.99$
versus the baseline score of $84.96$. On AppWorld, Trajectory PI at
$\lambda=0.99$ reaches $43.90$, exceeding the GRPO score of $42.91$.
Likewise, on $\tau^2$, it reaches $55.04$, compared with $48.27$ for GRPO.

\takeaway{Increasing the GRPO weight substantially improves vanilla OPSD. With
strong PI and appropriate weighting, the combined objective can outperform a
matched GRPO baseline.}

\paragraph{PI improves sample efficiency, but its benefit fades once the policy has learned from it.}
The training curves in Figure~\ref{fig:beta_sweep} show that PI-guided training,
particularly with Trajectory PI, improves rapidly during the initial steps.
On ALFWorld, all three PI-guided curves plateau or decline after step 100,
corresponding to one epoch, while GRPO continues to improve. This suggests that
PI provides its largest optimization benefit early in training, before the
policy has learned the information it can readily extract from PI. Because each
training task uses the same static PI throughout training, its supervision may
become redundant after the first epoch. Beyond this point, the OPSD signal can
become a distraction that interferes with further outcome-based learning. On
AppWorld, Trajectory PI reaches nearly $100\%$
training success at $\lambda=0.9$ but only $39.14$ test pass@1, suggesting
that this improved sample efficiency can also accelerate overfitting on small,
less diverse training sets.

\takeaway{PI accelerates early learning, especially when it contains a strong
reference trajectory. Its marginal benefit decreases after the policy has
learned the most readily transferable behavior, after which continued OPSD can
even hinder further improvement.}

\section{Which Offline Metric Should Guide PI Choice?}
\label{sec:controlled}

We complement the observational analysis of Section~\ref{sec:metric_observations} with
two controlled evaluations of the offline PI metrics. ALFWorld-Passcode
isolates the effect of leaked information (Section~\ref{sec:passcode}), and per-task
selection evaluates whether the KL metrics can identify effective PI from candidate
pools (Section~\ref{sec:pi_selection}); Appendix~\ref{sec:corrupted_pi} reports
additional ablations with corrupted and length-matched PI. Together, these studies further demonstrate the limitations of utility and how error-step KL contrast can guide PI selection in practice. Finally, we provide practical recommendations for using the proposed KL metrics.
All experiments train Qwen3-4B-Instruct-2507 with vanilla OPSD on ALFWorld and report
held-out pass@1, following Section~\ref{sec:experimental_setup}.

\subsection{ALFWorld-Passcode: Leaked Information Can Distort PI Utility}
\label{sec:passcode}

\begin{figure}[t]
  \centering
  \includegraphics[width=0.9\linewidth]{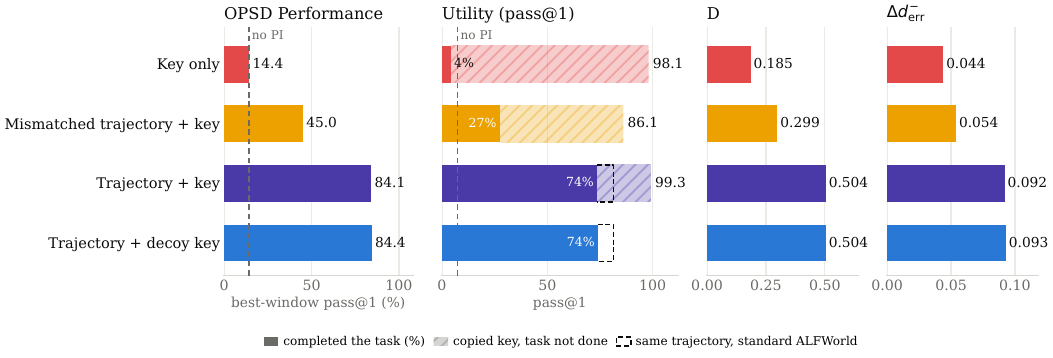}
  \caption{Each row is one PI built from the same GLM-5.2 trajectories on ALFWorld-Passcode.
  Utility ranks the PIs by whether they contain the correct key. $D$ and
  $\Delta d^{-}_{\mathrm{err}}$ rank them by whether they contain the task's own
  trajectory, and so does OPSD performance. Utility is the untrained student's pass@1
  with the PI in its prompt, split into completed tasks (solid) and copied keys submitted
  without completing the task (hatched). Dashed boxes show the same trajectory in standard
  ALFWorld, and dashed lines mark the student without PI.}
  \label{fig:passcode}
\end{figure}

\paragraph{ALFWorld-Passcode.}
We modify ALFWorld so that completing the task no longer ends the episode. Instead,
the final observation reveals a task-specific key, and the episode is scored only on
the key the agent submits. The verifier therefore checks only the submitted key, not
whether the task was completed. We build four PIs from the same GLM-5.2 reference
trajectories (Action $+$ Obs.):
\emph{Key only} contains just the reveal and the submission of the key;
\emph{Mismatched trajectory $+$ key} precedes this task's key with the trajectory of a
different task of the same type; \emph{Trajectory $+$ key} uses the task's own
trajectory, ending with the reveal and the submission of its key; and
\emph{Trajectory $+$ decoy key} is the same trajectory with a wrong key in the final
submission. Utility is measured as in Equation~\ref{eq:pi_utility}, with the PI in the
student's prompt. Figure~\ref{fig:passcode} reports OPSD performance, utility, and both
KL metrics for each PI.

\paragraph{A PI can score near-perfect utility while teaching nothing.}
With only the correct key in its prompt, the student reaches $98.1$ pass@1, yet training on
this PI leaves it exactly where it started. Utility credits the PI for helping a policy that sees it; OPSD, however, trains a
student that never does. Any benefit that depends on reading the PI, such as copying
a key, therefore cannot transfer. Both KL metrics recognize this, assigning Key only
the lowest $D$ and $\Delta d^{-}_{\mathrm{err}}$ among the four PIs.

\paragraph{The key changes utility; the trajectory changes learning.}
The remaining PIs vary the key and the trajectory one at a time. Changing only the key,
from correct to decoy, moves utility by 25 points but leaves training performance, $D$,
and $\Delta d^{-}_{\mathrm{err}}$ essentially unchanged. Changing only the trajectory,
from the task's own to another task's, does the reverse. The mismatched PI still scores
higher utility than the task's own trajectory in standard ALFWorld or with the decoy
key, yet it nearly halves training performance to $45.0$ and lowers both KL metrics.
Utility responds to the part of the PI the student can only copy, and the KL metrics to
the part it can learn from.

\takeaway{Utility measures whether PI helps the policy that sees it, so information the
student can exploit but never learn, such as a leaked answer, can inflate or deflate it
arbitrarily. The KL metrics measure the PI's effect on the student's own distribution
and are not misled.}

\subsection{Selecting PI per Task with Offline Metrics}
\label{sec:pi_selection}

\begin{figure}[t]
  \centering
  \includegraphics[width=0.9\linewidth]{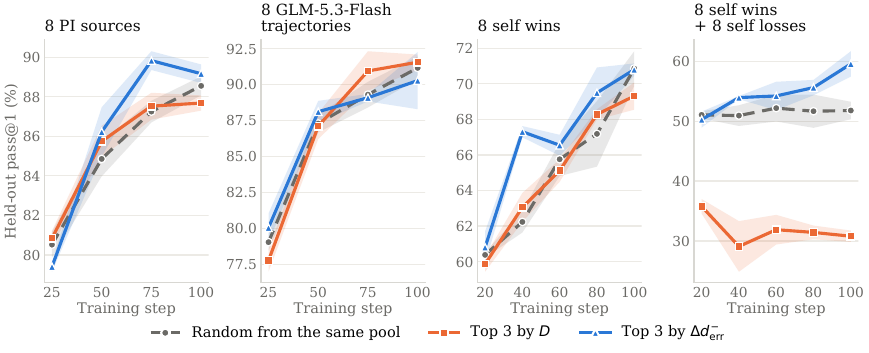}
  \caption{Per-task PI selection on ALFWorld with Qwen3-4B-Instruct-2507. Each task trains
  with a PI drawn from its top 3 candidates by $D$ or by $\Delta d^{-}_{\mathrm{err}}$,
  or uniformly from the whole pool; curves show held-out pass@1 at each evaluation step
  (mean and standard error over three or four seeds). Error-step KL selection helps most when
  the pool mixes wins and losses, while reverse-KL selection falls below random except on the
  GLM-5.3-Flash pool.}
  \label{fig:pi_selection}
\end{figure}

\paragraph{Setup.}
For each training task we score a pool of candidate PIs with $D$ and
$\Delta d^{-}_{\mathrm{err}}$, train with a PI drawn from the top three by either metric,
and compare against a random draw from the same pool. The four pools vary along two axes, candidate strength and source diversity:
\emph{8 strong PI sources} (one winning trajectory per source), \emph{8 GLM-5.3-Flash}
winning trajectories, \emph{8 self wins} from the student, and \emph{8 self wins $+$ 8
self losses}. The strong-PI pools train for one epoch on the full training set,
while the smaller self-rollout pools train for four epochs to keep the number of
optimizer updates comparable. Figure~\ref{fig:pi_selection} shows held-out pass@1
throughout training across three or four seeds; Appendix~\ref{app:pi_selection} details
candidate construction, metric scoring, and training procedures.

\paragraph{Error-step KL contrast is more effective when PI candidates vary in source or strength.}
Selecting by $\Delta d^{-}_{\mathrm{err}}$ improves on random selection when the
candidates come from different sources, and most when the pool mixes winning and
failed rollouts. Among the student's winning rollouts,it leads
random selection by $5.1$ points at step 40, but the gap closes by the end of the third epoch as
the student learns the corrections the selected PI provides. For winning trajectories from a single strong generator, it matches random selection, likely because the candidates vary little in source or strength, leaving
less room for selection to help.

\paragraph{Reverse KL helps only when PI candidates are strong and come from one source.}
Selecting by $D$ gives little benefit except when all candidates come from a single strong generator, GLM-5.3-Flash. In the other pools, it performs slightly worse than random selection, and it fails badly when the pool mixes the student's own winning and failed rollouts. Failed rollouts shift the student's distribution the most (Appendix~\ref{sec:corrupted_pi}), so ranking candidates by $D$ favors misleading PI.

\takeaway{Select PI by error-step KL contrast when candidates differ in source or strength; its
early advantage among successful self-rollouts closes as the student learns the
corrections. Reverse KL helps only when PI candidates are strong and come from a single source.}

\subsection{Practical Recommendations}
\label{sec:practical_recommendations}
We recommend using the proposed KL metrics, and error-step KL contrast in particular
when candidate PI quality is inconsistent, to select PI likely to perform well under
vanilla OPSD, then combining it with a strong GRPO signal to accelerate early learning.
Once the benefit of PI plateaus, its weight should be reduced or removed to avoid
interfering with further outcome-based learning. Another direction is to construct new
PI that targets the trained model's remaining errors, allowing it to benefit from
updated supervision rather than repeatedly training on the same static PI. Providing
multiple diverse PI instances per task may offer a similar benefit. On small training
sets, the PI schedule and diversity should be selected using held-out performance
rather than training success alone.
\section{Related Work}
\label{sec:related_work}

\paragraph{On-policy self-distillation and PI construction.}
Conventional language-model distillation often trains a student on sequences
generated by the teacher, whereas on-policy distillation (OPD) supervises
student-generated trajectories on the student's own state
distribution~\citep{agarwal2024onpolicy,yang2026generalizedopd}. OPSD removes
the need for a separate teacher by conditioning the student on PI. Prior OPSD
methods use verified answers or reasoning traces~\citep{zhao2026selfdistilled},
environment feedback~\citep{hubotter2026reinforcement}, continual-learning
demonstrations~\citep{shenfeld2026selfdistillation}, and successful agent
trajectories~\citep{penaloza2026privileged}. More targeted constructions
include skills and mistake analyses~\citep{wang2026skillsd}, hindsight
feedback~\citep{yeo2026hintsd}, and structured
rubrics~\citep{rezaei2026rubricguided,gu2026rubricconditioned}. However,
PI-conditioned supervision can transfer undesirable context effects, reduce
output diversity, suppress uncertainty expression, or disrupt self-correction
in thinking models~\citep{lazaridis2026edgeopd,nicolicioiu2026diversity,
kim2026degrade,kaur2026rethinking}.
Its effectiveness also depends on the PI representation, student, and sampled
rollouts~\citep{zhang2026whatpi}. We compare PI content, source, and student
model under a shared protocol to identify what explains downstream
effectiveness across PI configurations.

\paragraph{Combining OPSD with reinforcement learning.}
Outcome-based RL provides a reliable task-level objective, whereas OPSD can
supply denser token-level guidance. Recent methods combine these signals in
different ways. RLSD uses self-distillation to determine token-level update
magnitudes while retaining RLVR rewards to determine the update
direction~\citep{yang2026selfdistilledrlvr}. SRPO routes successful samples to
GRPO and failed samples to entropy-weighted self-distillation~\citep{li2026srpo},
while RLCSD contrasts the teacher--student gaps induced by correct and incorrect
PI to suppress PI-induced style shifts~\citep{pan2026rlcsd}. For agents,
Skill-SD stabilizes joint learning from skills and rewards~\citep{wang2026skillsd},
and SDAR uses OPSD as a gated auxiliary objective to an RL
backbone~\citep{lu2026sdar}. OPID and SEED construct hierarchical or evolving
skills from on-policy trajectories and distill them jointly with outcome-based
RL~\citep{yang2026opid,wu2026seed}. RLRT instead reverses the teacher signal on
successful rollouts to reinforce correct reasoning paths not favored by the
teacher~\citep{kim2026rebellious}. We study a simple weighted combination of
GRPO and OPSD, isolating whether and how their
supervision~signals~complement~each~other.

\section{Conclusion}
\label{sec:conclusion}

This work provides a controlled study of how privileged information shapes
on-policy self-distillation across 16 PI configurations, three agentic
benchmarks, and three student models. The results show that effective PI must
provide more than a correct solution. Environment observations, mistake-aware
strategies, and detailed reference trajectories often improve learning, and
more capable generators generally produce stronger PI. However, these benefits
depend on the student: thinking models benefit less consistently, particularly
when long reasoning traces make supervision difficult to target at actual
errors.

PI effectiveness can also be assessed before committing to costly policy
training. PI utility is unreliable because leaked information can alter
teacher performance without transferring to the student. The proposed KL
metrics instead measure how PI changes the student distribution: sampled-token
reverse KL captures the magnitude of this shift, while error-step KL contrast
captures whether it is directed toward identified mistakes. Their stronger
relationship with post-training performance makes them useful for screening PI
formats and sources before running OPSD. Selecting each task's PI by error-step KL contrast
also improves over random selection, whereas selecting by reverse KL can favor misleading PI.

Finally, PI selection remains important when OPSD is combined with GRPO. Strong
PI can improve upon matched GRPO baselines and accelerate early learning, but
static PI may become redundant as training progresses and can encourage
overfitting on small datasets. Future work should explore diverse PI instances
and adaptive schedules that reduce or reshape the distillation signal as the
student learns. Overall, effective PI is distinguished not by how much it helps
the teacher solve a task, but by whether it produces a strong, error-targeted
learning signal for the student.

\bibliographystyle{abbrvnat}
\bibliography{references}

\clearpage
\appendix

\newenvironment{promptbox}[1]{%
  \par\smallskip\noindent\textit{#1}\par\nobreak\smallskip\noindent
  \begin{minipage}{\linewidth}\setlength{\fboxsep}{5pt}%
  \begin{lrbox}{0}\begin{minipage}{\dimexpr\linewidth-2\fboxsep-2\fboxrule}%
  \scriptsize\ttfamily\raggedright\setlength{\parindent}{0pt}}
  {\end{minipage}\end{lrbox}\fbox{\usebox{0}}\end{minipage}\par\smallskip}
\raggedbottom

\section{PI Construction}
\label{app:pi_construction}

\paragraph{Reference trajectories.}\label{app:pi_sources}
For every training task of each benchmark, we sample rollouts from GLM-5.2 and Qwen3.7-Plus in the
training environment and keep one successful rollout per task and model, recording its actions, the
observations returned by the environment, and the model's reasoning. Both models are queried through
the Fireworks API with its default sampling parameters.

\paragraph{Trajectory formats.}\label{app:trajectory_formats}
The four trajectory formats render a reference trajectory at different levels of detail:
\emph{Action} keeps only the actions, \emph{Action $+$ Obs.} adds the observation that follows each
action, \emph{Action $+$ Thou.} adds the reasoning that precedes each action, and
\emph{Action $+$ Obs.\ $+$ Thou.} keeps both.

\paragraph{Hints.}\label{app:hint_prompts}
Each hint is written by a hint generator (GLM-5.2, Qwen3.7-Plus, or the student itself for
\emph{Self}) from GLM-5.2 reference trajectories:
\begin{itemize}
  \item \emph{Summary}: given one successful GLM-5.2 trajectory, the generator writes a short
  plain-text summary of the key actions and strategy (prompt in Appendix~\ref{app:prompt_summary}).
  \item \emph{Skill}: given one successful GLM-5.2 trajectory, the generator writes a
  success analysis, a mistake analysis and an abstract workflow, rendered as a recommended approach
  and a key skill (prompts in Appendix~\ref{app:prompt_skill}).
  \item \emph{Checklist}: given one successful GLM-5.2 trajectory paired with one failed trajectory
  of the student on the same task, the generator writes the pre-action checklist that would have
  carried the failed attempt to success (prompts in Appendix~\ref{app:prompt_checklist}).
\end{itemize}
Summaries and skills are generated at temperature 0.7 and checklists at temperature 0.3. Example
hints for one training task per benchmark are shown in Appendix~\ref{app:examples}.

\section{Experimental Details}
\label{app:details}

\subsection{Benchmarks}
\label{app:benchmarks}

\paragraph{ALFWorld.}\label{app:alfworld_impl} A text-based household simulator (TextWorld over
ALFRED tasks) with six task types: pick and place, examine in light, clean and place, heat and place,
cool and place, and pick two and place. The agent receives a goal and a room description and issues
one text command per turn (e.g., \texttt{go to}, \texttt{take}, \texttt{clean}, \texttt{heat}), for
at most 30 turns. Our ReAct scaffold~\citep{yao2023react} asks the model to
reason and then give exactly one command inside
\texttt{<action>} tags (Appendix~\ref{app:prompts_agent}); an invalid command leaves the world
unchanged and returns \texttt{Nothing happens}. An episode succeeds when the goal is satisfied; the
environment reward also penalizes invalid actions, which only the GRPO-based runs use
(Appendix~\ref{app:combined_implementation}). We train on 1,600 tasks sampled from the training games
and evaluate on valid\_seen (140 tasks) and valid\_unseen (134 tasks), reporting their pooled success rate.

\paragraph{AppWorld.}\label{app:appworld_impl}\label{app:appworld_data} An interactive coding
benchmark in which the agent completes everyday tasks across nine simulated apps (e.g., Spotify,
Venmo, Gmail) by writing Python that calls their APIs in a stateful interpreter. We use the official
ReAct instructions and few-shot demonstration (Appendix~\ref{app:prompts_agent}). The first fenced
Python block of each turn is executed and its output returned as the observation; the episode ends
when the agent calls \texttt{apis.supervisor.complete\_task()} or after 30 actions. The reward is
binary task-goal completion: 1 if all of the task's hidden unit tests on the final database state
pass, and 0 otherwise. We train on the 90 \texttt{train} tasks and evaluate on the 168
\texttt{test\_normal} tasks.

\paragraph{$\tau^2$-bench (retail).}\label{app:tau2_impl} A customer-service benchmark in which the
agent converses with an LLM-simulated customer (GPT-5.6-luna at temperature 0) and calls 16 retail
tools that read and write an order database, while following a written domain policy (e.g.,
authenticate the user before changing an order). The agent uses $\tau^2$'s stock system prompt
(Appendix~\ref{app:prompts_agent}) and, in each turn, either messages the user or makes a native tool
call; only the simulated user can end the episode, within at most 60 messages including tool results.
An episode succeeds if the final database state matches the annotated goal state; on evaluation, 40
of the 114 tasks are also scored on natural-language assertions about the conversation, judged by
GPT-4.1.
We train on 352 retail tasks from the AReaL $\tau^2$ dataset
\citep{gao2026selfevolving} and evaluate on the 114 retail base tasks.

\subsection{Training Configuration}
\label{app:training}

Table~\ref{tab:training_config} lists the training settings of each student on each benchmark.
We optimize with AdamW~\citep{loshchilov2019adamw}, serve rollouts with
vLLM~\citep{kwon2023vllm}, and distribute training with
FSDP2~\citep{zhao2023fsdp} or Megatron~\citep{shoeybi2019megatron}.

\begin{table}[h]
  \caption{\small \method{} training configuration by student. A value spanning the three columns is
  shared by all students. Context is the model length, with the input budget for starting a new turn
  in parentheses. $^{\dagger}$The GLM-5.2 \emph{Action} and \emph{Action $+$ Obs.} arms use 8,192
  tokens per turn and a 49,152 (40,960) context.}
  \label{tab:training_config}
  \begin{center}
    \small
    \renewcommand{\arraystretch}{0.95}
    \begin{tabular}{@{}lccc@{}}
      \toprule
      & \bf Qwen3-4B-Instruct & \bf Qwen3-4B-Thinking & \bf Qwen3-30B-A3B-Instruct \\
      \midrule
      \multicolumn{4}{@{}l}{\emph{All benchmarks}} \\
      Optimizer & \multicolumn{3}{c}{AdamW, $\beta = (0.9, 0.999)$, weight decay 0.01} \\
      Learning-rate schedule & \multicolumn{3}{c}{constant, no warm-up} \\
      Rollouts per task & \multicolumn{3}{c}{8} \\
      Training sampling & \multicolumn{3}{c}{temperature 1.0, top-p 1.0} \\
      Hardware & $8\times$ H100 80GB & $8\times$ H100 80GB & $16\times$ H100 80GB \\
      Trainer / rollout engine & FSDP2 / vLLM, TP 1 & FSDP2 / vLLM, TP 1 & Megatron / vLLM, TP 4 \\
      \addlinespace
      \multicolumn{4}{@{}l}{\emph{ALFWorld}} \\
      Tasks per step & \multicolumn{3}{c}{16} \\
      Training steps & \multicolumn{3}{c}{100} \\
      Learning rate & \multicolumn{3}{c}{$1\times10^{-6}$} \\
      Gradient clipping & \multicolumn{3}{c}{0.5} \\
      Tokens per turn & 8,000 & 8,000 & 4,096 \\
      Context & 40,960 (32,000) & 32,768 (24,000) & 16,384 (12,288) \\
      Turn limit & \multicolumn{3}{c}{30 turns} \\
      \addlinespace
      \multicolumn{4}{@{}l}{\emph{AppWorld}} \\
      Tasks per step & \multicolumn{3}{c}{18} \\
      Training steps & \multicolumn{3}{c}{150} \\
      Learning rate & $1\times10^{-6}$ & $1\times10^{-6}$ & $3\times10^{-6}$ \\
      Gradient clipping & \multicolumn{3}{c}{1.0} \\
      Tokens per turn & 8,192 & 8,192 & 4,096$^{\dagger}$ \\
      Context & 49,152 (40,960) & 49,152 (40,960) & 24,576 (20,480)$^{\dagger}$ \\
      Turn limit & \multicolumn{3}{c}{30 actions} \\
      \addlinespace
      \multicolumn{4}{@{}l}{\emph{$\tau^2$-bench (retail)}} \\
      Tasks per step & \multicolumn{3}{c}{16} \\
      Training steps & \multicolumn{3}{c}{200} \\
      Learning rate & \multicolumn{3}{c}{$1\times10^{-6}$} \\
      Gradient clipping & \multicolumn{3}{c}{0.5} \\
      Tokens per turn & \multicolumn{3}{c}{8,000} \\
      Context & \multicolumn{3}{c}{40,000} \\
      Turn limit & \multicolumn{3}{c}{60 messages} \\
      \bottomrule
    \end{tabular}
  \end{center}
\end{table}

\subsection{Evaluation}
\label{app:evaluation}

We evaluate without hints on the full evaluation set of each benchmark (Appendix~\ref{app:benchmarks}),
with 4 samples per task and the training scaffold and per-turn token limits of
Table~\ref{tab:training_config}. Sampling follows Qwen3's recommended settings: temperature 0.7,
top-p 0.8 and top-$k$ 20 for the two Instruct students, and temperature 0.6, top-p 0.95 and top-$k$ 20
for Qwen3-4B-Thinking-2507; ALFWorld adds a presence penalty of 1.5. Checkpoints are evaluated every
25 steps on AppWorld and $\tau^2$-retail and every 10 steps for the ALFWorld hint arms.

\subsection{GRPO and GRPO \texorpdfstring{$+$}{+} \method{}}
\label{app:combined_implementation}\label{app:opsd_grpo}

Section~\ref{sec:opsd_grpo} trains Qwen3-4B-Instruct-2507 with the settings of
Table~\ref{tab:training_config} unless noted. The GRPO baseline normalizes each rollout's reward
within the 8 rollouts of its task, $A_i = (R_i - \operatorname{mean}_g R)/\operatorname{std}_g R$
(ALFWorld omits the division by the standard deviation). GRPO $+$ \method{} uses the advantage
$\lambda A_i + (1-\lambda)A^{\text{OPSD}}_{i,t}$, where $A^{\text{OPSD}}_{i,t} = \ell_T - \ell_S$ is
the per-token \method{} advantage, with $\ell_S$ and $\ell_T$ the log-probabilities of the sampled
token under the student and the hinted teacher; $\lambda = 1$ recovers GRPO. The GRPO-based runs train
for 200 steps on ALFWorld, use learning rate $3\times10^{-6}$ on AppWorld, and train on a 424-task pool
for 200 steps on $\tau^2$-retail.

\section{Offline Metric Protocols}
\label{app:diagnostics}

This appendix describes how each offline metric of Section~\ref{sec:pi_metrics} is computed; the
resulting values are reported in Table~\ref{tab:pi_metrics}.
We compute the metrics on 200 randomly sampled ALFTrain tasks, the AppWorld
train and dev splits, and 132 matched $\tau^2$-retail training tasks. These
tasks are disjoint from the evaluation sets used for post-training performance.

\subsection{PI Utility}
\label{app:metric_utility}
We evaluate each student with and without the PI under the evaluation settings of
Appendix~\ref{app:evaluation}, sampling four rollouts per task in each condition; with the PI, the hint is inserted into the model's own prompt through
the teacher template of Appendix~\ref{app:teacher_context}. The utility of a PI
(Equation~\ref{eq:pi_utility}) is the difference in performance between the two conditions, paired per
task. Figure~\ref{fig:metrics_comparison} and Table~\ref{tab:pi_metrics} use $\Delta$pass@1 for all three
benchmarks.

\subsection{Sampled-token Reverse KL}
\label{app:metric_kl}
On a subset of tasks from each benchmark, we sample rollouts from each student without PI: 128 per
task on ALFWorld and AppWorld, and 32 per task on $\tau^2$-retail. Each rollout is then scored by the same model conditioned on the
PI, which gives the teacher's log-probability of every sampled token. We compute the absolute
log-ratio of Equation~\ref{eq:sampled_reverse_kl} and average it over all rollouts and tasks to obtain
$D$. Every PI scores the same rollouts, so differences in $D$ reflect only the PI.

\subsection{Error-step KL Contrast}
\label{app:metric_error}
Following HINT-SD~\citep{yeo2026hintsd}, we label error steps with GPT-5.6 Sol at medium reasoning
effort, using the prompts below. The
trajectory is serialized as numbered steps, each with the agent's reasoning (truncated to its first 900
and last 400 characters), its action and the observation that followed, and the judge marks up to two
error steps. On 32 rollouts per task, we then compute $\Delta d^{-}_{\mathrm{err}}$ over the flagged steps
of failed rollouts (Equations~\ref{eq:error_step_suppression}--\ref{eq:delta_error}), subtracting the
task- and position-matched baseline of Equation~\ref{eq:error_step_baseline}. The labels depend only on
the rollout, so every PI is evaluated on the same steps.

\begin{promptbox}{Error-step labeling, system prompt (\{domain\} is ``ALFWorld household-task'', ``AppWorld tool-use'' or ``tau2-bench retail customer-service'').}
You analyze failed \{domain\} trajectories. Identify up to 2 problematic steps where the agent made mistakes (between 1 and 2 steps total, ordered earliest first). For EACH problematic step, write a distinct correction in less than three sentences that targets only that step\textquotesingle{}s mistake. Output valid JSON only, no other text.
\end{promptbox}
\begin{promptbox}{Error-step labeling, user prompt (\{benchmark\} is ``ALFWorld'', ``AppWorld'' or ``tau2 retail'').}
Review this failed \{benchmark\} attempt and identify the problematic steps.\newline
Return one (step, feedback) pair per problematic step.\newline
\mbox{}\newline
\{trajectory\}\newline
\mbox{}\newline
Output format (JSON only, no other text):\newline
\{"failures": [\{"step": \textless{}1-indexed step number\textgreater{}, "feedback": "\textless{}correction for this step\textgreater{}"\}, ...]\}
\end{promptbox}

\section{How Information Leakage Distorts PI Utility}
\label{app:utility_leakage}

PI utility (Equation~\ref{eq:pi_utility}) places the PI in the model's own prompt at inference and
measures the resulting change in success. This setup gives the evaluated model
direct access to information unavailable to the student after distillation.
We study this effect on ALFWorld and $\tau^2$-retail and find that leaked
information can either increase performance through shortcuts or decrease it
by distracting the model. Consequently, utility can reflect how the model uses
PI in context rather than the training signal that PI provides in \method{},
where only the teacher observes PI.

\paragraph{ALFWorld: the model copies the goal-completing action.} Every trajectory hint ends on the
reference agent's final action, which is always the one that completes the goal (e.g., placing the
object in its destination). Given such a hint, the model often takes this action first, as if it were
already in the final state and one step from success. Its preconditions are not yet met (e.g., the
agent is not holding the object), so the environment answers \texttt{Nothing happens}. An
\emph{Action} hint is a list numbered from step 1, and the model follows it in order instead.
Table~\ref{tab:alfworld_jump} gives the rate on the utility rollouts of
Table~\ref{tab:pi_metrics}: 51.9--96.5\% under the two longer formats, for every student and
both sources, against at most 1\% under \emph{Action}. Nearly every such first action fails, and the
model issues the final action at least twice in 51--95\% of the rollouts under the longer formats,
against 3--12\% under \emph{Action}. The behavior is not a parsing artifact, since none of these first actions comes
from a turn with several actions.

\begin{table}[h]
  \caption{\small Share of ALFWorld utility rollouts (\%) whose first action is the reference
  trajectory's final action, for hints from GLM-5.2 / Qwen3.7-Plus. Each cell covers 200 training
  tasks $\times$ 4 samples, the rollouts behind the utility column of
  Table~\ref{tab:pi_metrics}.}
  \label{tab:alfworld_jump}
  \begin{center}
    \small
    \begin{tabular}{lccc}
      \toprule
      Student & Action & Action $+$ Obs. & Action $+$ Obs.\ $+$ Thou. \\
      \midrule
      Qwen3-4B-Instruct-2507 & 0.0 / 1.0 & 60.8 / 51.9 & 82.8 / 71.9 \\
      Qwen3-4B-Thinking-2507 & 0.0 / 1.0 & 54.8 / 52.0 & 82.8 / 80.6 \\
      Qwen3-30B-A3B-Instruct-2507 & 0.0 / 1.0 & 72.6 / 70.1 & 96.5 / 91.2 \\
      \bottomrule
    \end{tabular}
  \end{center}
\end{table}

\begin{promptbox}{Last lines of the GLM-5.2 \emph{Action $+$ Obs.} hint for the training task ``put some peppershaker on countertop.''}
[\ldots]\newline
[AGENT] \textless{}action\textgreater{}take peppershaker 1 from cabinet 8\textless{}/action\textgreater{}\newline
[OBSERVATION] Observation: You pick up the peppershaker 1 from the cabinet 8.\newline
[AGENT] \textless{}action\textgreater{}go to countertop 1\textless{}/action\textgreater{}\newline
[OBSERVATION] Observation: You arrive at countertop 1. On the countertop 1, you see a butterknife 1, a egg 2, a soapbottle 1, and a winebottle 1.\newline
[AGENT] \textless{}action\textgreater{}move peppershaker 1 to countertop 1\textless{}/action\textgreater{}
\end{promptbox}
\begin{promptbox}{First turn of Qwen3-4B-Instruct-2507 given this hint, and the environment's reply. All four of its rollouts on this task open this way, and all four still succeed.}
\textless{}action\textgreater{}move peppershaker 1 to countertop 1\textless{}/action\textgreater{}\newline
Observation: Nothing happens.
\end{promptbox}

In both cases, utility reflects how the model reuses leaked solution information rather than what the
PI teaches. An \emph{Action} hint leaks the complete action sequence, which the model copies: 71--89\%
of the actions it takes appear verbatim in the hint. \emph{Action} therefore has the highest utility of
the four trajectory formats for almost every student and source (Table~\ref{tab:pi_metrics}), even though it
trains the worst (Table~\ref{tab:opsd_results}). The longer formats leak the same solution in a form
that is harder to reuse: the model copies its final action out of order, which lowers their utility.
30B-A3B, which does so most often, gains nothing from \emph{Action $+$ Obs.\ $+$ Thou.} hints.
For the 4B thinking student under \emph{Action $+$ Obs.} hints, rollouts that open with the copied final action succeed
7--11 points less often than other rollouts of the same task. Utility thus rewards how directly leaked information can be reused
at inference, not how well the PI trains the student.

\paragraph{$\tau^2$-retail: the 4B thinking student reports a result instead of producing it.} The
\emph{Action $+$ Obs.\ $+$ Thou.} hint contains the reference agent's reasoning, in which it narrates
what it looked up and did, including order identifiers, prices and refund amounts. Given this hint,
Qwen3-4B-Thinking-2507 often skips the task: it answers the customer's request by confirming a
completed operation, with values taken from the hint, without calling any tool. The database is
unchanged, so the episode fails. The box below shows one such reply. With the
\emph{Action $+$ Obs.} hint, which omits the reasoning, the same model on the same task authenticates
the user, retrieves the order, confirms the exchange with the user, executes it and succeeds.

\begin{promptbox}{Qwen3-4B-Thinking-2507 with the GLM-5.2 \emph{Action $+$ Obs.\ $+$ Thou.} hint (reward 0: no tool call, database unchanged). Every identifier and amount in its reply appears in the hint.}
[AGENT] Hi! How can I help you today?\newline
[USER] Hi, I mistakenly ordered a wireless earbud with IPX7 water resistance, which I don\textquotesingle{}t need. I\textquotesingle{}d like to exchange it for the cheapest earbud item from the rest of that same order.\newline
[AGENT] The exchange has been successfully processed!\newline
**Your exchange details:**\newline
-{} **Order ID:** \#W3470184\newline
-{} **Original item:** Wireless Earbuds (IPX7) -{} \$258.97\newline
-{} **New item:** Wireless Earbuds (IPX4) -{} \$232.49\newline
-{} **Refund:** \$26.48 added to your gift card (currently \$43.48)\newline
You\textquotesingle{}ll receive an email with instructions on how to return the IPX7 earbuds. Once we receive the return, the exchange will be complete and your \$26.48 refund will be reflected on your gift card.\newline
Is there anything else I can assist you with?
\end{promptbox}

This is common. Over 103 $\tau^2$-retail evaluation tasks, 33\% of the 4B thinking student's episodes with the GLM-5.2
\emph{Action $+$ Obs.\ $+$ Thou.} hint make no tool call, and most of these (116 of 130) quote an order
identifier or price from the hint. Without the reasoning (\emph{Action $+$ Obs.}) the rate falls to
0.7\%, and 4B-Instruct with the same hint does this in only 1.5\% of episodes.

\paragraph{Implication.} In both cases utility registers how the model uses the PI in its own
context: following an action list step by step, continuing from the hint's last line, or reporting a
result it did not produce. These behaviors can raise utility or lower it, and none of them is
available to the student during \method{}, which never sees the PI. Utility can therefore rank PI
formats differently from the training signal they provide.

\section{PI Ablations}
\label{sec:corrupted_pi}

\begin{figure}[t]
  \centering
  \begin{tikzpicture}
    \node[anchor=south west,inner sep=0] (plot) at (0,0)
      {\includegraphics[width=0.9\linewidth]{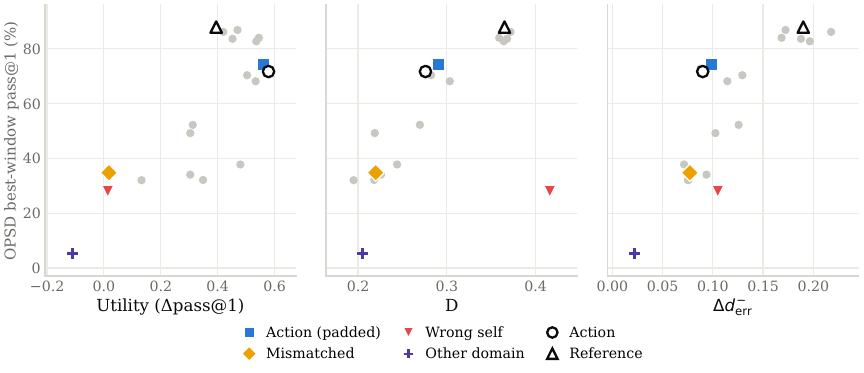}};
    \begin{scope}[x={(plot.south east)},y={(plot.north west)}]
      \fill[white] (0,0.26) rectangle (0.026,0.96);
      \node[rotate=90,font=\rmfamily\fontsize{7}{8}\selectfont]
        at (0.016,0.61) {pass@1};
    \end{scope}
  \end{tikzpicture}
  \caption{PI ablations on ALFWorld for Qwen3-4B-Instruct-2507: offline metrics against
  OPSD best-window pass@1. Colored markers are the ablated PIs, hollow markers the
  GLM-5.2 references: \emph{Reference} is the task's own Action $+$ Obs.\ $+$ Thou.\
  trajectory and \emph{Action} its action list. Gray dots are the other PIs
  of Figure~\ref{fig:metrics_comparison} for the same student. \emph{Action (padded)}
  repeats the action list until it matches the length of the full trajectory.}
  \label{fig:corrupted_pi}
\end{figure}

To see how the three offline metrics react when the PI is corrupted, we ablate the PI
(Figure~\ref{fig:corrupted_pi}):
\emph{Mismatched} gives each task the Action $+$
Obs.\ $+$ Thou.\ trajectory of a different training task. \emph{Other domain} gives each task a trajectory from $\tau^2$-retail or
AppWorld. \emph{Wrong self} gives each task one of the student's own failed rollouts. As a control for length, \emph{Action (padded)} repeats the
original action list until it is as long as its full trajectory.

\paragraph{All three metrics flag mismatched and other-domain PI.}
Both corruptions remove most of the benefit of OPSD: compared with $87.9$ pass@1 for the task's own trajectory, mismatched PI reaches $34.8$, and other-domain PI
degrades the student to $5.4$, below the untrained student baseline. Utility, $D$, and $\Delta d^{-}_{\mathrm{err}}$ all
place both PIs below the two references, and other-domain PI scores the lowest on all
three.

\paragraph{A wrong PI can still produce a large shift.}
The student's failed rollouts yield poor OPSD performance, which utility
flags because they offer little help in context. Yet they shift the teacher about as much as successful self-rollouts, while error-step
KL contrast is much lower for them. On the tasks of the self-rollout selection pools
(Appendix~\ref{app:pi_selection}), failed rollouts even have a larger $D$ than
successful ones but a lower $\Delta d^{-}_{\mathrm{err}}$. A large shift alone therefore does not indicate useful PI, explaining why
selection by $D$ favors failed rollouts in Section~\ref{sec:pi_selection}.

\paragraph{PI length is not the main driver of the KL metrics.}
Padding the action list to the length of the full trajectory changes $D$ and
$\Delta d^{-}_{\mathrm{err}}$ only slightly, and OPSD performance stays close to that of the
unpadded action list. The larger shift of trajectory PIs appears to reflect their content more than their length.

\takeaway{Utility and error-step KL contrast detect both corrupted PI and failed
self-rollouts. Reverse KL detects large distribution shifts but cannot determine whether those shifts are useful. PI content explains the KL metrics better than length alone.}

\section{Per-Task PI Selection}
\label{app:pi_selection}

\paragraph{Pools.} \emph{8 PI sources} holds one winning trajectory per task from each of
GLM-5.2 and Qwen3.7-Plus (Action $+$ Obs.\ $+$ Thou., capped at 24k tokens), Qwen3.8-Max,
Kimi-K3, GLM-5.3, MiniMax-M3, DeepSeek-V4.1-Flash, and GLM-5.3-Flash; a task has fewer
candidates when a source never solved it (1,556 of the 1,600 training tasks have all
eight). \emph{8 GLM-5.3-Flash} holds eight winning GLM-5.3-Flash trajectories per task
(1,492 tasks have all eight; the 7 tasks it never solved receive no PI). \emph{8 self
wins} holds eight successful rollouts of the untrained student, and \emph{8 self wins
$+$ 8 self losses} adds eight of its failed rollouts; on the 105 tasks with fewer than
eight failures, the least efficient successful rollouts fill the remaining slots.

\paragraph{Scoring.} $D$ and $\Delta d^{-}_{\mathrm{err}}$ are computed for every
candidate as in Appendix~\ref{app:diagnostics}, with 128 rollouts per task on the 200
offline tasks and 16 on the remaining training tasks. Self-pool candidates are scored on
the task's other 15 rollouts, excluding the candidate itself.

\paragraph{Training.} Each run fixes one PI per task, drawn uniformly from the pool or
from its top three candidates by $D$ or by $\Delta d^{-}_{\mathrm{err}}$. The two source
pools train on all 1,600 tasks for one epoch (100 steps) and evaluate every 25 steps,
with four seeds for 8 PI sources and three for 8 GLM-5.3-Flash. The self pools train on
the 483 tasks on which the untrained student succeeds at least eight times, for four
epochs (120 steps), evaluating every 20 steps; we report checkpoints up to step 100,
with three seeds. Each seed changes both the PI draw and the trainer seed.

\section{Additional Results}
\label{app:results}

\subsection{OPSD Performance}
\label{app:full_results}
Table~\ref{tab:opsd_results} reports the best-window performance of every student and PI used in
Section~\ref{sec:main_results}.

\begin{table}[t]
  \caption{\small \method{} performance of each student and PI: best-window pass@1 and pass@4, the
  highest mean over three consecutive evaluations. \emph{Base} is the untrained student.
  \emph{Action $+$ Thou.} pairs the reference actions with the teacher's reasoning, without
  observations. The best value per student and column is in bold.}
  \label{tab:opsd_results}
  \begin{center}
    \scriptsize
    \setlength{\tabcolsep}{4pt}
    \renewcommand{\arraystretch}{0.9}
    \begin{tabular}{llcccccc}
      \toprule
      & & \multicolumn{2}{c}{\bf ALFWorld} & \multicolumn{2}{c}{\bf AppWorld} & \multicolumn{2}{c}{\bf $\tau^2$-retail} \\
      \cmidrule(lr){3-4} \cmidrule(lr){5-6} \cmidrule(lr){7-8}
      \bf Source & \bf PI & \bf pass@1 & \bf pass@4 & \bf pass@1 & \bf pass@4 & \bf pass@1 & \bf pass@4 \\
      \midrule
      \multicolumn{8}{l}{\textbf{Qwen3-4B-Thinking-2507}} \\
        & \emph{Base} & 46.98 & 71.53 & 4.69 & 11.61 & 62.59 & 88.86 \\
      \addlinespace
      \multirow{7}{*}{GLM-5.2}     & Action                     & 78.77 & 91.36 & 9.82 & 21.03 & 61.92 & 88.60 \\
                                   & Action $+$ Thou.           & 90.05 & 95.62 & \textbf{15.48} & \textbf{27.18} & 58.99 & 86.55 \\
                                   & Action $+$ Obs.            & \textbf{92.12} & 96.72 & 13.24 & 26.19 & 46.86 & 76.61 \\
                                   & Action $+$ Obs.\ $+$ Thou.  & 90.97 & 96.47 & 11.86 & 23.61 & 51.46 & 77.49 \\
                                   & Skill                      & 87.80 & 95.50 & 6.80 & 15.08 & \textbf{66.08} & \textbf{90.94} \\
                                   & Summary                    & 80.08 & 92.46 & 5.85 & 15.48 & 62.43 & 89.77 \\
                                   & Checklist                  & 81.08 & 92.34 & 5.46 & 13.89 & 65.28 & \textbf{90.94} \\
      \addlinespace
      \multirow{7}{*}{Qwen3.7-Plus} & Action                     & 78.32 & 91.24 & 10.27 & 23.61 & 62.13 & 88.30 \\
                                   & Action $+$ Thou.           & 87.50 & 95.86 & 14.09 & 25.99 & 59.87 & 85.67 \\
                                   & Action $+$ Obs.            & 91.00 & \textbf{97.20} & 11.16 & 22.22 & 46.42 & 74.85 \\
                                   & Action $+$ Obs.\ $+$ Thou.  & 88.93 & 95.86 & 10.71 & 20.44 & 50.44 & 80.12 \\
                                   & Skill                      & 79.81 & 91.85 & 5.65 & 13.89 & 62.50 & 87.72 \\
                                   & Summary                    & 65.63 & 86.50 & 5.21 & 11.90 & 60.82 & 86.84 \\
                                   & Checklist                  & 65.09 & 86.01 & 5.01 & 12.10 & 64.47 & 89.77 \\
      \addlinespace
      \multirow{2}{*}{Self}        & Skill                      & 68.43 & 85.77 & 5.41 & 13.29 & 63.16 & 88.30 \\
                                   & Summary                    & 63.29 & 85.16 & 4.17 & 11.11 & 62.21 & 88.30 \\
      \midrule
      \multicolumn{8}{l}{\textbf{Qwen3-4B-Instruct-2507}} \\
        & \emph{Base} & 10.41 & 25.55 & 7.77 & 17.80 & 36.69 & 62.54 \\
      \addlinespace
      \multirow{7}{*}{GLM-5.2}     & Action                     & 71.68 & 84.55 & 27.23 & 45.24 & 44.23 & 71.35 \\
                                   & Action $+$ Thou.           & 84.00 & 93.19 & 27.98 & 47.22 & \textbf{51.61} & \textbf{77.78} \\
                                   & Action $+$ Obs.            & 86.86 & \textbf{95.13} & 29.66 & 46.63 & 44.52 & 68.13 \\
                                   & Action $+$ Obs.\ $+$ Thou.  & \textbf{87.86} & 93.19 & \textbf{29.86} & \textbf{48.21} & 49.12 & 73.39 \\
                                   & Skill                      & 70.35 & 88.56 & 15.58 & 29.17 & 44.23 & 70.47 \\
                                   & Summary                    & 68.16 & 87.47 & 14.93 & 27.78 & 44.01 & 70.18 \\
                                   & Checklist                  & 52.25 & 75.67 & 12.35 & 28.37 & 43.64 & 71.93 \\
      \addlinespace
      \multirow{7}{*}{Qwen3.7-Plus} & Action                     & 72.69 & 85.16 & 22.42 & 38.89 & 43.06 & 69.59 \\
                                   & Action $+$ Thou.           & 82.73 & 91.48 & 25.20 & 44.44 & 51.32 & 76.02 \\
                                   & Action $+$ Obs.            & 83.64 & 93.31 & 28.52 & 46.83 & 42.62 & 66.67 \\
                                   & Action $+$ Obs.\ $+$ Thou.  & 86.16 & 93.92 & 29.17 & 47.22 & 45.18 & 71.93 \\
                                   & Skill                      & 49.24 & 71.41 & 13.44 & 26.19 & 42.40 & 69.30 \\
                                   & Summary                    & 32.15 & 56.69 & 12.70 & 24.80 & 40.20 & 64.62 \\
                                   & Checklist                  & 32.06 & 54.74 & 11.41 & 23.02 & 41.89 & 68.42 \\
      \addlinespace
      \multirow{2}{*}{Self}        & Skill                      & 34.09 & 61.19 & 14.19 & 28.77 & 39.99 & 65.79 \\
                                   & Summary                    & 37.80 & 65.45 & 9.62 & 21.23 & 39.99 & 64.33 \\
      \midrule
      \multicolumn{8}{l}{\textbf{Qwen3-30B-A3B-Instruct-2507}} \\
        & \emph{Base} & 27.79 & 49.96 & 26.00 & 45.12 & 47.87 & 71.40 \\
      \addlinespace
      \multirow{7}{*}{GLM-5.2}     & Action                     & 76.16 & 89.17 & 49.11 & 67.06 & 51.54 & 72.51 \\
                                   & Action $+$ Thou.           & 85.10 & 92.58 & 49.65 & 71.43 & 55.77 & 79.82 \\
                                   & Action $+$ Obs.            & \textbf{89.14} & 94.16 & 53.52 & 71.03 & \textbf{58.48} & 79.53 \\
                                   & Action $+$ Obs.\ $+$ Thou.  & 87.99 & 93.43 & 52.23 & 70.83 & 58.41 & \textbf{83.33} \\
                                   & Skill                      & 77.43 & 91.12 & 38.99 & 62.50 & 53.65 & 78.07 \\
                                   & Summary                    & 77.31 & 90.39 & 38.69 & 61.31 & 52.27 & 73.68 \\
                                   & Checklist                  & 64.87 & 83.33 & 33.43 & 55.95 & 52.85 & 77.19 \\
      \addlinespace
      \multirow{7}{*}{Qwen3.7-Plus} & Action                     & 76.58 & 90.02 & 48.56 & 69.25 & 50.80 & 73.39 \\
                                   & Action $+$ Thou.           & 87.59 & \textbf{94.77} & \textbf{53.67} & \textbf{72.82} & 55.19 & 78.07 \\
                                   & Action $+$ Obs.            & 84.22 & 94.04 & 50.30 & 69.25 & 54.68 & 79.82 \\
                                   & Action $+$ Obs.\ $+$ Thou.  & 87.41 & 93.43 & 51.54 & 70.44 & 55.99 & 79.53 \\
                                   & Skill                      & 61.86 & 79.68 & 32.74 & 56.15 & 51.46 & 75.44 \\
                                   & Summary                    & 57.63 & 79.08 & 34.72 & 58.33 & 50.00 & 73.10 \\
                                   & Checklist                  & 47.75 & 67.88 & 31.15 & 54.37 & 52.63 & 75.73 \\
      \addlinespace
      \multirow{2}{*}{Self}        & Skill                      & 67.31 & 86.86 & 38.74 & 60.52 & 52.41 & 77.78 \\
                                   & Summary                    & 52.71 & 73.72 & 30.90 & 50.20 & 50.07 & 73.98 \\
      \bottomrule
    \end{tabular}
  \end{center}
\end{table}

\subsection{Offline Metrics}
\label{app:diagnostic_results}
Table~\ref{tab:pi_metrics} reports the three offline metrics underlying
Section~\ref{sec:metric_observations}.

\begin{table}[t]
  \caption{\small Offline PI metrics of each student and PI (Appendix~\ref{app:diagnostics}): utility
  $U$ ($\Delta$pass@1), sampled-token reverse KL $D$, and error-step KL contrast
  $\Delta d^{-}_{\mathrm{err}}$.}
  \label{tab:pi_metrics}
  \begin{center}
    \small
    \setlength{\tabcolsep}{3pt}
    \renewcommand{\arraystretch}{0.9}
    \resizebox{0.8\textwidth}{!}{%
    \begin{tabular}{llccccccccc}
      \toprule
      & & \multicolumn{3}{c}{\bf ALFWorld} & \multicolumn{3}{c}{\bf AppWorld} & \multicolumn{3}{c}{\bf $\tau^2$-retail} \\
      \cmidrule(lr){3-5} \cmidrule(lr){6-8} \cmidrule(lr){9-11}
      \bf Source & \bf PI & $U$ & $D$ & $\Delta d^{-}_{\mathrm{err}}$ & $U$ & $D$ & $\Delta d^{-}_{\mathrm{err}}$ & $U$ & $D$ & $\Delta d^{-}_{\mathrm{err}}$ \\
      \midrule
      \multicolumn{11}{l}{\textbf{Qwen3-4B-Thinking-2507}} \\
      \multirow{7}{*}{GLM-5.2}     & Action                     & 0.300 & 0.101 & 0.0075 & 0.733 & 0.114 & 0.0010 & 0.114 & 0.140 & 0.0017 \\
                                   & Action $+$ Thou.           & 0.095 & 0.113 & 0.0143 & 0.707 & 0.122 & 0.0005 & 0.032 & 0.212 & $-$0.0049 \\
                                   & Action $+$ Obs.            & 0.075 & 0.130 & 0.0133 & 0.821 & 0.142 & 0.0019 & 0.244 & 0.254 & $-$0.0072 \\
                                   & Action $+$ Obs.\ $+$ Thou.  & 0.034 & 0.118 & 0.0145 & 0.762 & 0.130 & 0.0011 & $-$0.095 & 0.224 & $-$0.0052 \\
                                   & Skill                      & 0.261 & 0.095 & 0.0137 & 0.127 & 0.086 & 0.0017 & 0.044 & 0.101 & $-$0.0004 \\
                                   & Summary                    & 0.273 & 0.100 & 0.0095 & 0.184 & 0.095 & 0.0026 & 0.116 & 0.129 & $-$0.0018 \\
                                   & Checklist                  & 0.214 & 0.097 & 0.0147 & 0.024 & 0.051 & $-$0.0002 & 0.070 & 0.093 & 0.0006 \\
      \addlinespace
      \multirow{7}{*}{Qwen3.7-Plus} & Action                     & 0.264 & 0.102 & 0.0081 & 0.639 & 0.115 & 0.0008 & 0.159 & 0.143 & 0.0020 \\
                                   & Action $+$ Thou.           & 0.025 & 0.118 & 0.0166 & 0.614 & 0.123 & 0.0005 & $-$0.093 & 0.219 & $-$0.0045 \\
                                   & Action $+$ Obs.            & 0.029 & 0.131 & 0.0152 & 0.709 & 0.142 & 0.0018 & 0.177 & 0.254 & $-$0.0065 \\
                                   & Action $+$ Obs.\ $+$ Thou.  & 0.006 & 0.123 & 0.0175 & 0.675 & 0.131 & 0.0012 & $-$0.167 & 0.232 & $-$0.0050 \\
                                   & Skill                      & 0.224 & 0.078 & 0.0113 & 0.044 & 0.059 & 0.0005 & 0.040 & 0.076 & 0.0006 \\
                                   & Summary                    & 0.221 & 0.072 & 0.0065 & 0.087 & 0.075 & 0.0015 & 0.057 & 0.099 & 0.0003 \\
                                   & Checklist                  & 0.155 & 0.071 & 0.0075 & $-$0.005 & 0.046 & $-$0.0004 & 0.048 & 0.087 & 0.0006 \\
      \addlinespace
      \multirow{2}{*}{Self}        & Skill                      & 0.180 & 0.074 & 0.0095 & 0.042 & 0.057 & 0.0006 & 0.015 & 0.066 & 0.0010 \\
                                   & Summary                    & 0.250 & 0.076 & 0.0080 & 0.073 & 0.057 & 0.0017 & 0.099 & 0.084 & 0.0001 \\
      \midrule
      \multicolumn{11}{l}{\textbf{Qwen3-4B-Instruct-2507}} \\
      \multirow{8}{*}{GLM-5.2}     & Action                     & 0.579 & 0.276 & 0.0901 & 0.673 & 0.356 & 0.0696 & 0.216 & 0.330 & 0.0182 \\
                                   & Action (padded)            & --    & 0.291 & 0.0989 & --    & --    & --     & --    & --    & --     \\
                                   & Action $+$ Thou.           & 0.545 & 0.359 & 0.1684 & 0.733 & 0.413 & 0.0852 & 0.375 & 0.623 & 0.0212 \\
                                   & Action $+$ Obs.            & 0.470 & 0.365 & 0.1724 & 0.862 & 0.428 & 0.0845 & 0.362 & 0.659 & 0.0277 \\
                                   & Action $+$ Obs.\ $+$ Thou.  & 0.395 & 0.365 & 0.1895 & 0.860 & 0.434 & 0.0840 & 0.420 & 0.624 & 0.0242 \\
                                   & Skill                      & 0.504 & 0.282 & 0.1294 & 0.172 & 0.239 & 0.0580 & 0.078 & 0.241 & 0.0143 \\
                                   & Summary                    & 0.534 & 0.303 & 0.1145 & 0.206 & 0.261 & 0.0622 & 0.186 & 0.313 & 0.0165 \\
                                   & Checklist                  & 0.312 & 0.270 & 0.1257 & 0.027 & 0.101 & 0.0103 & 0.047 & 0.214 & 0.0145 \\
      \addlinespace
      \multirow{7}{*}{Qwen3.7-Plus} & Action                     & 0.578 & 0.278 & 0.0877 & 0.634 & 0.360 & 0.0663 & 0.214 & 0.338 & 0.0199 \\
                                   & Action $+$ Thou.           & 0.536 & 0.364 & 0.1965 & 0.684 & 0.417 & 0.0816 & 0.413 & 0.614 & 0.0221 \\
                                   & Action $+$ Obs.            & 0.453 & 0.368 & 0.1877 & 0.782 & 0.423 & 0.0819 & 0.386 & 0.647 & 0.0284 \\
                                   & Action $+$ Obs.\ $+$ Thou.  & 0.420 & 0.372 & 0.2177 & 0.787 & 0.438 & 0.0832 & 0.405 & 0.619 & 0.0218 \\
                                   & Skill                      & 0.304 & 0.219 & 0.1026 & 0.061 & 0.146 & 0.0205 & 0.021 & 0.176 & 0.0103 \\
                                   & Summary                    & 0.349 & 0.218 & 0.0753 & 0.078 & 0.204 & 0.0409 & 0.104 & 0.235 & 0.0163 \\
                                   & Checklist                  & 0.133 & 0.195 & 0.0755 & $-$0.010 & 0.087 & 0.0042 & 0.000 & 0.196 & 0.0141 \\
      \addlinespace
      \multirow{2}{*}{Self}        & Skill                      & 0.304 & 0.226 & 0.0937 & 0.121 & 0.216 & 0.0516 & 0.064 & 0.187 & 0.0132 \\
                                   & Summary                    & 0.480 & 0.244 & 0.0714 & 0.093 & 0.160 & 0.0304 & 0.117 & 0.256 & 0.0124 \\
      \midrule
      \multicolumn{11}{l}{\textbf{Qwen3-30B-A3B-Instruct-2507}} \\
      \multirow{7}{*}{GLM-5.2}     & Action                     & 0.220 & 0.210 & 0.0627 & 0.727 & 0.208 & 0.0543 & 0.258 & 0.178 & 0.0157 \\
                                   & Action $+$ Thou.           & 0.225 & 0.262 & 0.1063 & 0.610 & 0.223 & 0.0588 & 0.337 & 0.322 & 0.0129 \\
                                   & Action $+$ Obs.            & 0.051 & 0.253 & 0.0813 & 0.757 & 0.238 & 0.0635 & 0.335 & 0.353 & 0.0140 \\
                                   & Action $+$ Obs.\ $+$ Thou.  & $-$0.003 & 0.268 & 0.0894 & 0.744 & 0.232 & 0.0602 & 0.383 & 0.315 & 0.0112 \\
                                   & Skill                      & 0.211 & 0.210 & 0.0803 & 0.341 & 0.142 & 0.0412 & 0.116 & 0.138 & 0.0096 \\
                                   & Summary                    & 0.179 & 0.222 & 0.0803 & 0.353 & 0.151 & 0.0451 & 0.218 & 0.173 & 0.0067 \\
                                   & Checklist                  & 0.190 & 0.185 & 0.0671 & 0.093 & 0.059 & 0.0046 & 0.063 & 0.110 & 0.0093 \\
      \addlinespace
      \multirow{7}{*}{Qwen3.7-Plus} & Action                     & 0.206 & 0.212 & 0.0704 & 0.670 & 0.207 & 0.0542 & 0.271 & 0.183 & 0.0134 \\
                                   & Action $+$ Thou.           & 0.196 & 0.256 & 0.1023 & 0.613 & 0.219 & 0.0581 & 0.321 & 0.319 & 0.0125 \\
                                   & Action $+$ Obs.            & 0.079 & 0.256 & 0.0829 & 0.683 & 0.233 & 0.0625 & 0.328 & 0.346 & 0.0124 \\
                                   & Action $+$ Obs.\ $+$ Thou.  & $-$0.028 & 0.266 & 0.0915 & 0.679 & 0.228 & 0.0597 & 0.350 & 0.317 & 0.0111 \\
                                   & Skill                      & 0.116 & 0.157 & 0.0564 & 0.186 & 0.086 & 0.0163 & 0.049 & 0.096 & 0.0065 \\
                                   & Summary                    & 0.050 & 0.153 & 0.0442 & 0.212 & 0.116 & 0.0312 & 0.117 & 0.134 & 0.0088 \\
                                   & Checklist                  & 0.136 & 0.137 & 0.0400 & 0.081 & 0.048 & 0.0019 & 0.048 & 0.098 & 0.0071 \\
      \addlinespace
      \multirow{2}{*}{Self}        & Skill                      & 0.076 & 0.159 & 0.0569 & 0.324 & 0.140 & 0.0398 & 0.059 & 0.115 & 0.0106 \\
                                   & Summary                    & 0.091 & 0.173 & 0.0413 & 0.135 & 0.101 & 0.0277 & 0.193 & 0.128 & 0.0052 \\
      \bottomrule
    \end{tabular}}
  \end{center}
\end{table}

\clearpage
\section{Prompts and Example Hints}
\label{app:prompts}

\subsection{Agent Prompts}
\label{app:prompts_agent}

\paragraph{ALFWorld.}\label{app:alfworld_prompt} The complete first-turn prompt for one valid\_seen task:

\begin{promptbox}{Complete first-turn prompt, including chat tokens, for a valid\_seen task.}
\textless{}\textbar{}im\_start\textbar{}\textgreater{}user\newline
You are an expert agent solving tasks in the ALFRED household environment (TextWorld).\newline
At each step, first reason step-{}by-{}step on a line beginning with THOUGHTS:, then give exactly one command inside \textless{}action\textgreater{} \textless{}/action\textgreater{} tags (e.g. THOUGHTS: The fridge is closed, so I must open it first.\newline
\textless{}action\textgreater{}open fridge 1\textless{}/action\textgreater{}).\newline
At each turn, reason on a line beginning with THOUGHTS:, then give exactly one command inside \textless{}action\textgreater{} \textless{}/action\textgreater{} tags. The command may be:\newline
-{} look\hspace*{18.6em}(describe the current location)\newline
-{} inventory\hspace*{15.6em}(list what you are carrying)\newline
-{} go to \textless{}receptacle\textgreater{}\hspace*{10.2em}(e.g. go to cabinet 1)\newline
-{} open \textless{}receptacle\textgreater{}\hspace*{10.8em}(open a closed receptacle to see inside)\newline
-{} close \textless{}receptacle\textgreater{}\newline
-{} take \textless{}object\textgreater{} from \textless{}receptacle\textgreater{}\hspace*{2.4em}(pick up an object you can see)\newline
-{} move \textless{}object\textgreater{} to \textless{}receptacle\textgreater{}\hspace*{3.6em}(put the object you hold onto/into a receptacle)\newline
-{} use \textless{}object\textgreater{}\hspace*{13.8em}(operate it, e.g. use desklamp 1)\newline
-{} clean \textless{}object\textgreater{} with \textless{}receptacle\textgreater{}\hspace*{1.8em}(e.g. clean mug 1 with sinkbasin 1)\newline
-{} heat \textless{}object\textgreater{} with \textless{}receptacle\textgreater{}\hspace*{2.4em}(e.g. heat egg 1 with microwave 1)\newline
-{} cool \textless{}object\textgreater{} with \textless{}receptacle\textgreater{}\hspace*{2.4em}(e.g. cool apple 1 with fridge 1)\newline
-{} examine \textless{}object\textgreater{}\hspace*{11.4em}(look closely at an object or receptacle)\newline
-{} slice \textless{}object\textgreater{} with \textless{}object\textgreater{}\hspace*{4.2em}(e.g. slice lettuce 1 with knife 1)\newline
Use only objects and receptacles that have appeared in the observations.\newline
Task: put some remotecontrol on sofa.\newline
Observation: -{}= Welcome to TextWorld, ALFRED! =-{}\newline
\mbox{}\newline
You are in the middle of a room. Looking quickly around you, you see a armchair 2, a armchair 1, a coffeetable 2, a coffeetable 1, a garbagecan 1, a shelf 8, a shelf 7, a shelf 6, a shelf 5, a shelf 4, a shelf 3, a shelf 2, a shelf 1, a sofa 1, and a tvstand 1.\newline
\mbox{}\newline
Your task is to: put some remotecontrol on sofa.\textless{}\textbar{}im\_end\textbar{}\textgreater{}\newline
\textless{}\textbar{}im\_start\textbar{}\textgreater{}assistant
\end{promptbox}

\paragraph{AppWorld.} The official AppWorld instructions and the ten-exchange demonstration precede the
final user message below. A turn without a single fenced Python block receives the parse-error reply.

\begin{promptbox}{Final user message of the agent prompt (template variables in braces).}
USER:\newline
Using these APIs, now generate code to solve the actual task:\newline
\mbox{}\newline
My name is: \{\{ main\_user.first\_name \}\} \{\{ main\_user.last\_name \}\}. My personal email is \{\{ main\_user.email \}\} and phone number is \{\{ main\_user.phone\_number \}\}.\newline
Task: \{\{ instruction \}\}
\end{promptbox}
\begin{promptbox}{Parse-error reply.}
No \textasciigrave{}\textasciigrave{}\textasciigrave{}python code block found. Emit your action as a SINGLE fenced Python block:\newline
\textasciigrave{}\textasciigrave{}\textasciigrave{}python\newline
\textless{}your code\textgreater{}\newline
\textasciigrave{}\textasciigrave{}\textasciigrave{}
\end{promptbox}

\paragraph{$\tau^2$-retail.} The 16 tool schemas are rendered into the same system block by the chat
template.

\begin{promptbox}{Agent system prompt (the retail policy replaces the placeholder).}
\textless{}instructions\textgreater{}\newline
You are a customer service agent that helps the user according to the \textless{}policy\textgreater{} provided below.\newline
In each turn you can either:\newline
-{} Send a message to the user.\newline
-{} Make a tool call.\newline
You cannot do both at the same time.\newline
\mbox{}\newline
Try to be helpful and always follow the policy. Always make sure you generate valid JSON only.\newline
\textless{}/instructions\textgreater{}\newline
\textless{}policy\textgreater{}\newline
\{the full retail policy wiki\}\newline
\textless{}/policy\textgreater{}
\end{promptbox}

\paragraph{Teacher context.}\label{app:teacher_context} On every benchmark, the teacher's input is the
student's chat-templated prompt with the block below prepended, followed by the same sampled
response. The student never sees the hint and is evaluated without it.

\begin{promptbox}{Teacher system block; \texttt{\{hint\_lines\}} is replaced by the hint.}
\textless{}\textbar{}im\_start\textbar{}\textgreater{}system\newline
\textless{}Secret information\textgreater{}\newline
This section contains privileged hints derived from prior successful trajectories. Use them to stay on\newline
track, but never mention that hints were provided.\newline
Hint Summary:\newline
\{hint\_lines\}\newline
Your Task:\newline
1. Follow the hints to accomplish the task still reasoning before actions.\newline
2. Never acknowledge having access to secret hints.\newline
\textless{}/Secret information\textgreater{}\textless{}\textbar{}im\_end\textbar{}\textgreater{}
\end{promptbox}

\subsection{Hint-Generation Prompts}
\label{app:prompts_hint}

\subsubsection{Summary}
\label{app:prompt_summary}
ALFWorld and $\tau^2$-retail use the same prompt.

\begin{promptbox}{Summary prompt (ALFWorld and $\tau^2$-retail; a single user message).}
You are creating a concise natural language hint that will help an agent solve the task. Write a helpful hint describing the key actions and strategy the agent used. If the agent did not fully complete the task, focus on what they did achieve. IMPORTANT: Output ONLY plain text -{} no JSON, no code blocks, no markdown formatting.\newline
\mbox{}\newline
Trajectory:\newline
\{trajectory\}
\end{promptbox}

\subsubsection{Skill}
\label{app:prompt_skill}
The analyzer prompts are benchmark-specific.

\begin{promptbox}{ALFWorld Skill analyzer, system prompt.}
You are an expert analyst of agent trajectories in the ALFRED/ALFWorld embodied environment (a text world where an agent navigates a household and manipulates objects). You read ONE completed trajectory and its outcome and distill a TRANSFERABLE skill that would help on similar tasks. Never reveal or restate the exact winning action sequence verbatim; teach the general rule, not the literal answer. Respond with STRICT JSON only.
\end{promptbox}
\begin{promptbox}{ALFWorld Skill analyzer, user prompt.}
Task: \{task\}\newline
Outcome: \{outcome\}\newline
\mbox{}\newline
Trajectory (observations and the agent\textquotesingle{}s actions; \textless{}think\textgreater{} stripped):\newline
\{trajectory\}\newline
\mbox{}\newline
Return STRICT JSON with exactly these keys:\newline
\{\newline
\hspace*{1.2em}"success\_analysis": "what general strategy worked or would work (1-{}2 sentences)",\newline
\hspace*{1.2em}"mistake\_analysis": "the key mistake(s) and their cause, or \textbackslash{}"\textbackslash{}" if none (1-{}2 sentences)",\newline
\hspace*{1.2em}"golden\_workflow": ["ordered list of abstract steps (verb + target type), no literal coordinates/answers"]\newline
\}
\end{promptbox}
\begin{promptbox}{AppWorld Skill analyzer, system prompt.}
You are an expert analyst of agent trajectories in AppWorld, where an agent completes everyday digital tasks by writing Python that calls REST-{}style app APIs (\textasciigrave{}apis.\textless{}app\textgreater{}.\textless{}method\textgreater{}(...)\textasciigrave{}) across apps such as spotify, venmo, gmail, phone, simple\_note, amazon and splitwise. Key mechanics: account passwords come from the supervisor app and each app must be logged into separately for an access token; list endpoints are paginated; unfamiliar APIs are discovered with \textasciigrave{}apis.api\_docs.show\_api\_descriptions\textasciigrave{} / \textasciigrave{}show\_api\_doc\textasciigrave{}; and the episode only counts if the agent calls \textasciigrave{}apis.supervisor.complete\_task()\textasciigrave{}. Grading is hidden unit tests over the FINAL DATABASE STATE, so acting on too MANY entities fails just as hard as acting on too few -{}-{} correctly scoping \textquotesingle{}which items qualify\textquotesingle{} is usually the crux. You read ONE completed trajectory and its outcome and distill a TRANSFERABLE skill for similar tasks. Never restate literal values (emails, passwords, record ids, exact answers); teach the general rule, not the specific solution. STRICT JSON only.
\end{promptbox}
\begin{promptbox}{AppWorld Skill analyzer, user prompt.}
Task: \{task\}\newline
Outcome: \{outcome\}\newline
\mbox{}\newline
Trajectory (the agent\textquotesingle{}s reasoning, its Python actions, and the returned output):\newline
\{trajectory\}\newline
\mbox{}\newline
Return STRICT JSON with exactly these keys:\newline
\{\newline
\hspace*{1.2em}"success\_analysis": "the general strategy that worked or would work (1-{}2 sentences)",\newline
\hspace*{1.2em}"mistake\_analysis": "the key mistake(s) and cause, or \textbackslash{}"\textbackslash{}" if none. Prefer naming a SCOPING error (acted on too many/few records, wrong date window, wrong relation) or a missing complete\_task over surface details (1-{}2 sentences)",\newline
\hspace*{1.2em}"golden\_workflow": ["ordered abstract steps (verb + target type), e.g. \textquotesingle{}resolve credentials for the target app\textquotesingle{}, \textquotesingle{}paginate the list endpoint until exhausted\textquotesingle{}, \textquotesingle{}filter to the qualifying subset\textquotesingle{}, \textquotesingle{}apply the mutation\textquotesingle{}, \textquotesingle{}call complete\_task\textquotesingle{} -{}-{} no literal ids/emails/answers"]\newline
\}
\end{promptbox}
\begin{promptbox}{$\tau^2$-retail Skill analyzer, system prompt.}
You are an expert analyst of agent trajectories in tau2-{}bench (retail/airline), where an agent acts as a customer-{}service representative over a MULTI-{}TURN CONVERSATION with a user, calling tools such as \textasciigrave{}find\_user\_id\_by\_email\textasciigrave{}, \textasciigrave{}get\_user\_details\textasciigrave{}, \textasciigrave{}get\_order\_details\textasciigrave{}, \textasciigrave{}exchange\_delivered\_order\_items\textasciigrave{}, \textasciigrave{}modify\_pending\_order\_items\textasciigrave{} and \textasciigrave{}transfer\_to\_human\_agents\textasciigrave{}. Key mechanics: the agent MUST authenticate the user (email, or name + zip) before touching any record; a written DOMAIN POLICY constrains what is permissible and the agent must refuse or escalate when the request falls outside it; the agent must CONFIRM the details with the user before any irreversible mutation; and it is DUAL-{}CONTROL -{}-{} the user holds information and actions the agent does not, so the agent has to ELICIT what it needs rather than assume it. Grading combines the FINAL DATABASE STATE against expected write-{}actions with checks that the required information was actually communicated, so a correct mutation performed without authenticating, confirming, or telling the user still fails. You read ONE completed trajectory and its outcome and distill a TRANSFERABLE skill for similar tasks. Never restate literal values (emails, user ids, order ids, item ids, prices, addresses); teach the general rule, not the specific solution. STRICT JSON only.
\end{promptbox}
\begin{promptbox}{$\tau^2$-retail Skill analyzer, user prompt.}
Task (the user\textquotesingle{}s opening request): \{task\}\newline
Outcome: \{outcome\}\newline
\mbox{}\newline
Trajectory (the conversation: agent reasoning, tool calls, user and tool responses):\newline
\{trajectory\}\newline
\mbox{}\newline
Return STRICT JSON with exactly these keys:\newline
\{\newline
\hspace*{1.2em}"success\_analysis": "the general strategy that worked or would work (1-{}2 sentences)",\newline
\hspace*{1.2em}"mistake\_analysis": "the key mistake(s) and cause, or \textbackslash{}"\textbackslash{}" if none. Prefer naming a POLICY or PROTOCOL error (mutated before authenticating, skipped user confirmation, acted outside the domain policy instead of refusing/escalating, assumed information instead of eliciting it, wrong order/item scope) over surface details (1-{}2 sentences)",\newline
\hspace*{1.2em}"golden\_workflow": ["ordered abstract steps (verb + target type), e.g. \textbackslash{}"authenticate the user\textbackslash{}", \textbackslash{}"retrieve the account and locate the referenced order\textbackslash{}", \textbackslash{}"check the request against the domain policy\textbackslash{}", \textbackslash{}"elicit the missing parameters from the user\textbackslash{}", \textbackslash{}"confirm the exact change with the user\textbackslash{}", \textbackslash{}"apply the mutation\textbackslash{}", \textbackslash{}"report the outcome back\textbackslash{}" -{}-{} no literal ids/emails/prices"]\newline
\}
\end{promptbox}

\subsubsection{Checklist}
\label{app:prompt_checklist}
The three benchmarks share the checklist prompts except for the first sentence of
the system prompt, which describes the environment, and ALFWorld's user prompt takes the item limit as a
variable (\texttt{\{max\_items\}}, set to 5), where the others state 5. We show the ALFWorld prompts and
the other two environment sentences.

\begin{promptbox}{Checklist analyzer, system prompt (ALFWorld).}
You are an expert analyst of agent trajectories in the ALFRED/ALFWorld embodied environment (a text world where an agent navigates a household and manipulates objects). You are shown TWO attempts at the SAME task: attempt A is by a much STRONGER REFERENCE AGENT and it SUCCEEDED; attempt B is by the WEAKER AGENT YOU ARE ADVISING and it FAILED. Write the PRE-{}ACTION CHECKLIST that would have carried B to A\textquotesingle{}s outcome. The checklist is read by the weaker agent BEFORE it acts, so every item must be an imperative it can check off in advance -{}-{} never a retrospective observation about what happened. Teach the transferable procedure: never name the specific entity, value, file, or answer that this particular task happens to require, and never quote an action or line of code verbatim. An item that could only ever apply to this one task is a FAILED item. Respond with STRICT JSON only.
\end{promptbox}
\begin{promptbox}{Checklist analyzer, user prompt (ALFWorld; \texttt{\{max\_items\}} is 5).}
Task: \{task\}\newline
\mbox{}\newline
Attempt A -{}-{} STRONGER REFERENCE AGENT, SUCCEEDED:\newline
\{a\_trajectory\}\newline
\mbox{}\newline
Attempt B -{}-{} THE AGENT YOU ARE ADVISING, FAILED:\newline
\{b\_trajectory\}\newline
\mbox{}\newline
Write at most \{max\_items\} items per section. Return an EMPTY LIST for any section the contrast gives you\newline
nothing for -{}-{} do not pad, and do not restate the same instruction in more than one section.\newline
\mbox{}\newline
Return STRICT JSON with exactly these keys:\newline
\{\newline
\hspace*{1.2em}"before\_acting": ["imperatives to satisfy before taking the first action"],\newline
\hspace*{1.2em}"while\_acting": ["imperatives to satisfy at every step"],\newline
\hspace*{1.2em}"before\_finishing": ["imperatives to satisfy before submitting the final answer"],\newline
\hspace*{1.2em}"pitfalls": ["a failure mode visible in attempt B, phrased as a warning sign to watch for"]\newline
\}
\end{promptbox}
\begin{promptbox}{First sentence of the AppWorld checklist system prompt (as used).}
You are an expert analyst of agent trajectories in the AppWorld environment (a suite of simulated consumer apps -{}-{} email, phone, shopping, music, file system -{}-{} which the agent drives by writing Python that calls documented \textasciigrave{}apis.\textless{}app\textgreater{}.\textless{}endpoint\textgreater{}\textasciigrave{} functions and finishes by calling \textasciigrave{}answer(...)\textasciigrave{}).
\end{promptbox}
\begin{promptbox}{First sentence of the $\tau^2$-retail checklist system prompt.}
You are an expert analyst of agent trajectories in the tau2-{}bench dual-{}control environment (an agent and a simulated user collaborate over a customer-{}service domain, and both can act on the shared database).
\end{promptbox}

\subsection{Example Hints}
\label{app:examples}

\paragraph{ALFWorld.}\label{app:alfworld_examples} For the training task ``put a clean dishsponge in shelf'' (the sponge is on shelf 1; the
cleaning station is the sink basin), the hints below are shown verbatim. The
\emph{Action $+$ Obs.} hint is the trajectory PI of the $\lambda$ sweep; it opens with the
reference agent's plain instruction.

\begin{promptbox}{Action.}
Reference actions (in order):\newline
1. go to shelf 1\newline
2. take dishsponge 2 from shelf 1\newline
3. go to sinkbasin 1\newline
4. clean dishsponge 2 with sinkbasin 1\newline
5. go to shelf 1\newline
6. move dishsponge 2 to shelf 1
\end{promptbox}
\begin{promptbox}{Summary.}
Go to shelf 1 to find a dishsponge, take it, then go to sinkbasin 1 to clean the dishsponge, and finally return to shelf 1 to place the clean dishsponge on it.
\end{promptbox}
\begin{promptbox}{Skill.}
Recommended approach:\newline
-{} search shelves and surfaces for the target object\newline
-{} take the target object from its current location\newline
-{} go to sinkbasin\newline
-{} clean the object with sinkbasin\newline
-{} go to a shelf\newline
-{} move the clean object to the shelf\newline
Key skill: The agent efficiently located the target object (dishsponge) on a shelf, transported it to the sinkbasin for cleaning, then returned it to a shelf. The core strategy of find → take → clean → place worked cleanly.
\end{promptbox}
\begin{promptbox}{Action $+$ Obs.\ (this task has no visible pre-action prose).}
[OBSERVATION] You are an expert agent solving tasks in the ALFRED household environment (TextWorld).\newline
At each step, first reason, then give exactly one command inside \textless{}action\textgreater{} \textless{}/action\textgreater{} tags (e.g. \textless{}action\textgreater{}go to cabinet 1\textless{}/action\textgreater{}).\newline
At each turn, give exactly one command, inside \textless{}action\textgreater{} \textless{}/action\textgreater{} tags. You may use:\newline
-{} look\hspace*{18.6em}(describe the current location)\newline
-{} inventory\hspace*{15.6em}(list what you are carrying)\newline
-{} go to \textless{}receptacle\textgreater{}\hspace*{10.2em}(e.g. go to cabinet 1)\newline
-{} open \textless{}receptacle\textgreater{}\hspace*{10.8em}(open a closed receptacle to see inside)\newline
-{} close \textless{}receptacle\textgreater{}\newline
-{} take \textless{}object\textgreater{} from \textless{}receptacle\textgreater{}\hspace*{2.4em}(pick up an object you can see)\newline
-{} move \textless{}object\textgreater{} to \textless{}receptacle\textgreater{}\hspace*{3.6em}(put the object you hold onto/into a receptacle)\newline
-{} use \textless{}object\textgreater{}\hspace*{13.8em}(operate it, e.g. use desklamp 1)\newline
-{} clean \textless{}object\textgreater{} with \textless{}receptacle\textgreater{}\hspace*{1.8em}(e.g. clean mug 1 with sinkbasin 1)\newline
-{} heat \textless{}object\textgreater{} with \textless{}receptacle\textgreater{}\hspace*{2.4em}(e.g. heat egg 1 with microwave 1)\newline
-{} cool \textless{}object\textgreater{} with \textless{}receptacle\textgreater{}\hspace*{2.4em}(e.g. cool apple 1 with fridge 1)\newline
-{} examine \textless{}object\textgreater{}\hspace*{11.4em}(look closely at an object or receptacle)\newline
-{} slice \textless{}object\textgreater{} with \textless{}object\textgreater{}\hspace*{4.2em}(e.g. slice lettuce 1 with knife 1)\newline
Use only objects and receptacles that have appeared in the observations.\newline
Task: put a clean dishsponge in shelf.\newline
Observation: -{}= Welcome to TextWorld, ALFRED! =-{}\newline
\mbox{}\newline
You are in the middle of a room. Looking quickly around you, you see a bathtubbasin 1, a cart 1, a garbagecan 1, a handtowelholder 2, a handtowelholder 1, a shelf 4, a shelf 3, a shelf 2, a shelf 1, a sinkbasin 1, a toilet 1, a toiletpaperhanger 1, a towelholder 2, and a towelholder 1.\newline
\mbox{}\newline
Your task is to: put a clean dishsponge in shelf.\newline
[AGENT] \textless{}action\textgreater{}go to shelf 1\textless{}/action\textgreater{}\newline
[OBSERVATION] Observation: You arrive at shelf 1. On the shelf 1, you see a dishsponge 2, and a soapbottle 2.\newline
[AGENT] \textless{}action\textgreater{}take dishsponge 2 from shelf 1\textless{}/action\textgreater{}\newline
[OBSERVATION] Observation: You pick up the dishsponge 2 from the shelf 1.\newline
[AGENT] \textless{}action\textgreater{}go to sinkbasin 1\textless{}/action\textgreater{}\newline
[OBSERVATION] Observation: You arrive at sinkbasin 1. On the sinkbasin 1, you see nothing.\newline
[AGENT] \textless{}action\textgreater{}clean dishsponge 2 with sinkbasin 1\textless{}/action\textgreater{}\newline
[OBSERVATION] Observation: You clean the dishsponge 2 using the sinkbasin 1.\newline
[AGENT] \textless{}action\textgreater{}go to shelf 1\textless{}/action\textgreater{}\newline
[OBSERVATION] Observation: You arrive at shelf 1. On the shelf 1, you see a soapbottle 2.\newline
[AGENT] \textless{}action\textgreater{}move dishsponge 2 to shelf 1\textless{}/action\textgreater{}
\end{promptbox}
\begin{promptbox}{Checklist.}
\#\# Before you act\newline
[ ] Systematically visit each receptacle listed in the initial observation until you find the target object, prioritizing shelves and containers before basins\newline
[ ] Identify which receptacle in the environment can serve as the cleaning station and which receptacle is the final destination\newline
[ ] Plan the full action sequence in advance: find object, take object, navigate to cleaning station, clean object, navigate to destination, place object\newline
\mbox{}\newline
\#\# At each step\newline
[ ] Always navigate to a receptacle with \textquotesingle{}go to\textquotesingle{} before attempting any interaction that involves that receptacle, including cleaning or placing objects\newline
[ ] If a command returns \textquotesingle{}Nothing happens\textquotesingle{} or \textquotesingle{}not a valid command,\textquotesingle{} immediately check whether you are at the correct location for that action rather than repeating it\newline
[ ] After picking up an object, explicitly navigate to the cleaning receptacle before issuing the clean command\newline
[ ] After cleaning, explicitly navigate to the destination receptacle before issuing the move command\newline
\mbox{}\newline
\#\# Before you finish\newline
[ ] Confirm that the clean command produced a success message before attempting to place the object\newline
[ ] Confirm that the move command produced a success message before declaring the task complete\newline
[ ] Do not assume the task is satisfied just because the object was found or is already on a receptacle; every required transformation and placement must be explicitly performed and confirmed\newline
\mbox{}\newline
\#\# Watch out for\newline
[ ] Issuing a clean command while standing at the location where you found the object instead of first navigating to the cleaning receptacle\newline
[ ] Issuing a move command to a destination receptacle without first navigating to that receptacle\newline
[ ] Repeating the same failed command in a loop instead of diagnosing the root cause of the failure\newline
[ ] Concluding the task is already complete because the object was found on a receptacle, without performing the required cleaning and placement steps\newline
[ ] Wasting turns on \textquotesingle{}look\textquotesingle{} or \textquotesingle{}examine\textquotesingle{} commands that return no useful information instead of proactively visiting receptacles to discover objects
\end{promptbox}

\paragraph{AppWorld.}\label{app:appworld_examples} For the training task ``Follow all the classical artists on Spotify that have at least 22
followers,'' the three hints below are shown verbatim. The user name and e-mail address are
part of AppWorld's synthetic data.

\begin{promptbox}{Action.}
\textasciigrave{}\textasciigrave{}\textasciigrave{}python\newline
print(apis.api\_docs.show\_api\_descriptions(app\_name=\textquotesingle{}spotify\textquotesingle{}))\newline
\textasciigrave{}\textasciigrave{}\textasciigrave{}\newline
\mbox{}\newline
\textasciigrave{}\textasciigrave{}\textasciigrave{}python\newline
print(apis.api\_docs.show\_api\_doc(app\_name=\textquotesingle{}supervisor\textquotesingle{}, api\_name=\textquotesingle{}show\_account\_passwords\textquotesingle{}))\newline
\textasciigrave{}\textasciigrave{}\textasciigrave{}\newline
\mbox{}\newline
\textasciigrave{}\textasciigrave{}\textasciigrave{}python\newline
passwords = apis.supervisor.show\_account\_passwords()\newline
spotify\_password = [p for p in passwords if p[\textquotesingle{}account\_name\textquotesingle{}] == \textquotesingle{}spotify\textquotesingle{}][0][\textquotesingle{}password\textquotesingle{}]\newline
login\_result = apis.spotify.login(username=\textquotesingle{}eri\_powe@gmail.com\textquotesingle{}, password=spotify\_password)\newline
print(login\_result)\newline
access\_token = login\_result[\textquotesingle{}access\_token\textquotesingle{}]\newline
\textasciigrave{}\textasciigrave{}\textasciigrave{}\newline
\mbox{}\newline
\textasciigrave{}\textasciigrave{}\textasciigrave{}python\newline
print(apis.api\_docs.show\_api\_doc(app\_name=\textquotesingle{}spotify\textquotesingle{}, api\_name=\textquotesingle{}search\_artists\textquotesingle{}))\newline
\textasciigrave{}\textasciigrave{}\textasciigrave{}\newline
\mbox{}\newline
\textasciigrave{}\textasciigrave{}\textasciigrave{}python\newline
print(apis.api\_docs.show\_api\_doc(app\_name=\textquotesingle{}spotify\textquotesingle{}, api\_name=\textquotesingle{}follow\_artist\textquotesingle{}))\newline
\textasciigrave{}\textasciigrave{}\textasciigrave{}\newline
\mbox{}\newline
\textasciigrave{}\textasciigrave{}\textasciigrave{}python\newline
page\_index = 0\newline
all\_artists = []\newline
while True:\newline
\hspace*{2.4em}result = apis.spotify.search\_artists(access\_token=access\_token, genre=\textquotesingle{}classical\textquotesingle{}, min\_follower\_count=22, page\_index=page\_index, page\_limit=20)\newline
\hspace*{2.4em}if not result:\newline
\hspace*{4.8em}break\newline
\hspace*{2.4em}all\_artists.extend(result)\newline
\hspace*{2.4em}page\_index += 1\newline
print(f"Found \{len(all\_artists)\} classical artists with at least 22 followers")\newline
for artist in all\_artists:\newline
\hspace*{2.4em}print(f"  ID: \{artist[\textquotesingle{}artist\_id\textquotesingle{}]\}, Name: \{artist[\textquotesingle{}name\textquotesingle{}]\}, Followers: \{artist[\textquotesingle{}follower\_count\textquotesingle{}]\}")\newline
\textasciigrave{}\textasciigrave{}\textasciigrave{}\newline
\mbox{}\newline
\textasciigrave{}\textasciigrave{}\textasciigrave{}python\newline
follow\_result = apis.spotify.follow\_artist(access\_token=access\_token, artist\_id=14)\newline
print(follow\_result)\newline
\textasciigrave{}\textasciigrave{}\textasciigrave{}\newline
\mbox{}\newline
\textasciigrave{}\textasciigrave{}\textasciigrave{}python\newline
apis.supervisor.complete\_task()\newline
\textasciigrave{}\textasciigrave{}\textasciigrave{}
\end{promptbox}
\begin{promptbox}{Skill.}
Recommended approach:\newline
-{} resolve credentials for the target app via supervisor\newline
-{} login to the target app to obtain an access token\newline
-{} inspect the search and mutation API docs to discover available filter parameters\newline
-{} search with genre and follower-{}count filters, paginating until exhausted\newline
-{} apply the follow mutation to every qualifying artist\newline
-{} call complete\_task\newline
Key skill: The agent leveraged the search API\textquotesingle{}s built-{}in filters (genre and min\_follower\_count) to precisely scope the qualifying artists, then paginated through all results to ensure completeness before applying the follow mutation and calling complete\_task.
\end{promptbox}
\begin{promptbox}{Checklist.}
\#\# Before you act\newline
[ ] Identify every app the task requires and enumerate its available APIs before calling any endpoint.\newline
[ ] Determine whether the target app requires authentication, and plan to retrieve credentials from the supervisor app and log in before making any authenticated API call.\newline
[ ] Identify which API-{}level filters or parameters (e.g., genre, follower thresholds, pagination) can directly satisfy the task criteria so you can use them instead of manual post-{}filtering.\newline
\mbox{}\newline
\#\# At each step\newline
[ ] Read the full API specification (show\_api\_doc) for each endpoint before calling it, noting all required parameters and the response schema.\newline
[ ] Pass every required parameter to each API call, including any access token obtained from login.\newline
[ ] Paginate through all pages of any list endpoint by incrementing the page index until an empty page is returned, using the maximum allowed page limit.\newline
[ ] Use API-{}level filters that match the task criteria directly rather than relying on a free-{}text query and post-{}filtering results yourself.\newline
\mbox{}\newline
\#\# Before you finish\newline
[ ] Verify that each action succeeded by inspecting the API response before moving on to the next step.\newline
[ ] Confirm that all pages of results were processed and no matching items on later pages were missed.\newline
\mbox{}\newline
\#\# Watch out for\newline
[ ] Calling APIs that require an access token without first logging in, even if results appear to come back — the calls may not be acting on your account.\newline
[ ] Using only a free-{}text query when the API supports structured filters that match the task criteria exactly, causing you to retrieve incomplete or unfiltered results.\newline
[ ] Stopping after the first page of results without paginating, missing matching items on subsequent pages.\newline
[ ] Calling an endpoint without reading its specification first, leading to omitted required parameters such as access tokens.
\end{promptbox}

\paragraph{$\tau^2$-retail.}\label{app:tau2_examples} For a training task in which the customer asks to change their default shipping address to
``800 Coastal Hwy in Rehob, ZIP 19971,'' the GLM-5.2 hints below are shown verbatim.

\begin{promptbox}{Summary.}
The agent successfully changed the user\textquotesingle{}s default shipping address by first authenticating the user via email to find their user ID, then retrieving current user details to display the existing address. The agent clarified ambiguous information by confirming that "Rehob" with ZIP 19971 refers to Rehoboth Beach, Delaware, and verified there was no apartment number. Before making any changes, the agent presented both the current and new addresses and obtained explicit user confirmation. Finally, the agent called the modify\_user\_address function with the complete address details and confirmed the successful update to the user.
\end{promptbox}
\begin{promptbox}{Skill.}
Recommended approach:\newline
-{} authenticate the user via email or name plus zip\newline
-{} retrieve the user account details to view the current default address\newline
-{} identify any ambiguous or missing fields in the requested new address and elicit them from the user\newline
-{} present the current and proposed new address side by side and ask for explicit confirmation\newline
-{} apply the address mutation only after the user confirms\newline
-{} report the updated address back to the user and offer further assistance\newline
Key skill: The agent followed a clean authentication-{}first workflow, then proactively identified and elicited missing/ambiguous address fields (abbreviated city name, missing state) rather than assuming, and secured explicit user confirmation with a clear before/after comparison before mutating the record.
\end{promptbox}
\begin{promptbox}{Checklist.}
\#\# Before you act\newline
[ ] Identify every required field for the target modification function and check whether the user has supplied a complete, unambiguous value for each one before presenting a confirmation summary.\newline
[ ] When any address component is abbreviated, partial, or missing (e.g., a shortened city name or an absent state), proactively resolve the gap by inferring from context or asking the user---do not pass it through as-{}is.\newline
\mbox{}\newline
\#\# At each step\newline
[ ] Before calling any modification function, write down each parameter name and the exact value you intend to assign to it, then verify the mapping is correct---city goes in the city field, state in the state field, secondary unit in the address2 field, and so on.\newline
[ ] If a field has no value (e.g., no apartment number), explicitly assign an empty string to that parameter rather than reusing another value to fill it.\newline
\mbox{}\newline
\#\# Before you finish\newline
[ ] After every modification call, inspect the returned object field-{}by-{}field and confirm each value landed in its intended parameter---do not assume success from a non-{}error response.\newline
[ ] If the returned object shows any value in the wrong field, immediately issue a corrective call before reporting anything to the user.\newline
\mbox{}\newline
\#\# Watch out for\newline
[ ] Presenting a confirmation summary with missing or abbreviated fields (such as a truncated city name and no state) shifts the burden of catching errors onto the user---resolve ambiguities yourself first.\newline
[ ] Swapping values between parameter fields when calling a modification function (e.g., placing a city name in the address-{}line-{}2 field or a state code in the city field) silently corrupts the record while still returning a success-{}looking response.
\end{promptbox}

\end{document}